\pdfoutput=1

\documentclass[a4paper]{article}

\usepackage[utf8]{inputenc}
\usepackage[margin=1in]{geometry}
\usepackage[dvipsnames]{xcolor}
\usepackage[colorlinks]{hyperref}
\usepackage[colorinlistoftodos]{todonotes}
\usepackage[noend]{algpseudocode}
\usepackage{amssymb, amsmath, amsthm, algorithm, authblk, bm, booktabs, lineno, tikz}
\usepackage{hyperref}

\usepackage{amsmath}
\usepackage{amssymb}
\usepackage{amsthm}
\usepackage{mathpazo}
\usepackage{txfonts}

\DeclareMathOperator*{\argmin}{arg\,min}

\usepackage{graphicx}
\usepackage{pifont}
\usepackage{fancyvrb}
\usepackage{pdfpages}
\usepackage{float}
\usepackage{stmaryrd}
\usepackage{booktabs} 
\usepackage{enumitem}

\usepackage{tikz}
\usetikzlibrary{bayesnet,fit,positioning,decorations.shapes}
\tikzset{> = {triangle 45}}

\theoremstyle{plain}
\newtheorem{proposition}{Proposition}

\theoremstyle{definition}
\newtheorem{definition}{Definition}

\theoremstyle{remark}

\title{Learning Tractable Probabilistic Models\\ for Moral Responsibility and Blame\footnote{Published in \emph{Data Mining and Knowledge Discovery}. DOI: \href{https://doi.org/10.1007/s10618-020-00726-4}{10.1007/s10618-020-00726-4}.}}
\author[1]{Lewis Hammond\footnote{Work conducted while at the University of Edinburgh. Correspondence to \href{mailto:lewis.hammond@cs.ox.ac.uk}{\texttt{lewis.hammond@cs.ox.ac.uk}}.}}
\author[2,3]{Vaishak Belle}
\affil[1]{University of Oxford}
\affil[2]{University of Edinburgh} 
\affil[3]{Alan Turing Institute}
\date{}

\begin{document}
	
\maketitle 

\begin{abstract}
Moral responsibility is a major concern in autonomous systems, with applications ranging from self-driving cars to kidney exchanges. Although there have been recent attempts to formalise responsibility and blame, among similar notions, the problem of learning within these formalisms has been unaddressed. From the viewpoint of such systems, the urgent questions are: (a) How can models of moral scenarios and blameworthiness be extracted and learnt automatically from data? (b) How can judgements be computed effectively and efficiently, given the split-second decision points faced by some systems? By building on constrained tractable probabilistic learning, we propose and implement a hybrid (between data-driven and rule-based methods) learning framework for inducing models of such scenarios automatically from data and reasoning tractably from them. We report on experiments that compare our system with human judgement in three illustrative domains: lung cancer staging, teamwork management, and trolley problems.
\end{abstract}

\section{Introduction}
Moral responsibility is a major concern in autonomous systems. In applications ranging from self-driving cars to kidney exchanges \cite{conitzer2017moral}, contextualising and enabling judgements of morality and blame is becoming a difficult challenge, owing in part to the philosophically vexing nature of these notions. In the infamous trolley problem \cite{thomson1985trolley}, for example, a putative agent encounters a runaway trolley headed towards five  individuals who are unable to escape the trolley's path. Their death is certain if the trolley were to collide  with them. The agent, however, can save them by diverting the trolley to a side track by means of a switch, but at the cost of the death of another individual, who happens to be on this latter track. While one would hope that in practice the situations encountered by, say, self-driving cars would not involve such extreme choices (many of which may already be covered under the law or other regulations \cite{etzioni2017incorporating}), in our view it is crucial that AI systems act in line with human values and preferences. Imbuing such systems with the ability to reason about moral value, blame, and intentionality is one possible step towards this goal.

Moral reasoning has been actively studied by philosophers, lawyers, and psychologists for many decades. Within the context of interactions between humans and autonomous systems, the notion of blameworthiness has been argued as being critical to effective collaboration, decision-making, and to our thoughts about morality in general \cite{kim2006should,groom2010critic}. In many frameworks, especially the limited number that are quantitative, a definition of responsibility that is based on causality and counterfactuals has been argued to be particularly appealing. For example, Malle et al. \cite{malle2014theory} argue that for blame to emerge, an agent must be perceived as the \textit{cause} of a negative event. Similarly, Chockler and Halpern \cite{chockler2004responsibility} extend the definition of causality given by Halpern and Pearl \cite{halpern2005causes} to account for the \textit{degree} of responsibility (as opposed to an `all or nothing' definition). However, in each of these frameworks and definitions the problem of learning has been unaddressed. Instead, the theories are motivated and instantiated by carefully constructed examples designed by the expert, and so are not necessarily viable in large-scale applications. Indeed, problematic situations encountered by autonomous systems are likely to be in a high-dimensional setting, with large numbers of latent variables capturing the low-level aspects of the application domain, and potentially requiring fast judgements. Thus, the urgent questions are: 
\begin{itemize}[leftmargin=6mm]
	\item[(a)] \emph{How can models of moral scenarios and blameworthiness be extracted and learnt automatically from data?} 
	\item[(b)] \emph{How can judgements be computed effectively and efficiently, given the split-second decision points faced by some systems?}
\end{itemize} 
In this work, we propose and implement a \emph{hybrid learning framework for inducing models of moral scenarios and blameworthiness automatically from data, and reasoning tractably from them}. To the best of our knowledge, this is the first of such proposals. We remark that \emph{we do not motivate any new definitions for moral responsibility, but show how an existing formal framework} \cite{halpern2018towards} \emph{can be embedded in our learning framework}. We suspect it should be possible to analogously embed  other definitions from the literature too, and refer the reader to \cite{halpern2018towards,malle2014theory} for a discussion of alternative logics and frameworks. 

The demands on our learning framework are two-fold. First, it must support the efficient learning of probabilities. Second, it must be able to compute decisions (i.e., probabilistic queries)  efficiently. To address these  challenges in general, the  tractable learning paradigm has recently emerged  \cite{poon2011sum,choi2015tractable,kisa2014probabilistic}, which can induce both high- and low-treewidth graphical models with latent variables. In this sense, they realise a deep probabilistic architecture. Most significantly,   conditional or marginal distributions can be computed in time linear in the size of the model.  
We discuss how the class of tractable models considered in \cite{kisa2014probabilistic} turn out to be particularly appropriate for the task at hand. Overall, our primary contributions within this work can be grouped into three main areas: theoretical details of an embedding between our chosen framework and class of model, including complexity results; a fully implemented demonstration version of our system \cite{code}; and a series of experimental results, together with discussion of the more philosophical aspects of our work. 

We begin in Section \ref{background} with an introduction to the particular framework for moral responsibility and class of model that we use in our framework, along with a simple, illustrative example of each. In Section \ref{embedding} we present our embedding between this framework and model, before providing a series of complexity results (section \ref{complexityresults}). Details of our implementation are given in Section \ref{implementation}, with full documentation to be included alongside our code. We then report on experiments (section \ref{experiments}) regarding the alignment between automated and human judgements of moral decision-making in three illustrative domains: lung cancer staging, teamwork management, and trolley problems. Finally, in Section \ref{discussion} we discuss some of the more philosophical issues surrounding our work, its motivation, and its potential applications, before concluding with a look at related work and directions for future research (sections \ref{related} and \ref{conclusion} respectively).

\section{Preliminaries}
\label{background}

In this section we discuss an existing formal framework around which we develop a learning framework. In particular we build on the causality-based definition from Halpern and Kleiman-Weiner \cite{halpern2018towards}, henceforth HK, discussed in more detail below. We also provide a brief technical introduction to our model of choice, Probabilistic Sentential Decision Diagrams (PSDDs) \cite{kisa2014probabilistic}, and a brief example in Subsection \ref{example} illustrating the use of each.

\subsection{Blameworthiness}
\label{blame}

 In order to avoid ambiguity, we follow the authors of HK by using the word \textit{blameworthiness} to capture an important part of what can more broadly be described as moral responsibility, and consider a set of definitions taken directly from their original work, with slight changes in notation for the sake of clarity and conciseness. In HK, environments are modelled in terms of variables and structural equations relating their values \cite{halpern2005causes}. More formally, the variables are partitioned into \textit{exogenous} variables $\mathcal{X}$ external to the model in question, and \textit{endogenous} variables $\mathcal{V}$ that are internal to the model and whose values are determined by those of the exogenous variables and some subset of the already determined endogenous variables. A range function $\mathcal{R}$ maps every variable to the set of possible values it may take. We abuse notation slightly by writing $\mathcal{R}(\mathcal{Y})$ instead of $\times_{Y \in \mathcal{Y}}\mathcal{R}(Y)$ for a set of variables $\mathcal{Y}$. In any model, there exists one structural equation $F_V : \mathcal{R}(\mathcal{X} \cup \mathcal{V} \setminus \{V\}) \rightarrow \mathcal{R}(V)$ for each $V \in \mathcal{V}$.

\theoremstyle{definition}
\begin{definition}
	A \textbf{causal model} $M$ is a pair $(\mathcal{S, F})$ where $\mathcal{S}$ is a \textbf{signature} $(\mathcal{X, V, R})$ and $\mathcal{F}$ is a set of \textbf{modifiable structural equations} $\{F_V : V \in \mathcal{V}\}$. A \textbf{causal setting} is a pair $(M, \textbf{X})$, where $\textbf{X} \in \mathcal{R}(\mathcal{X})$ is a called a \textbf{context}.
\end{definition}

In general we denote an assignment of values to variables in a set $\mathcal{Y}$ as $\textbf{Y}$. Following HK, we restrict our considerations to \textit{recursive} models $M$, in which, given a context $\textbf{X}$, the values of all variables in $\mathcal{V}$ are uniquely determined. We denote this unique valuation by $\textbf{V}_{(M, \textbf{X})}$.

\theoremstyle{definition}
\begin{definition}
    \label{prim}
	A \textbf{primitive event} is an equation of the form $V = v$ for some $V \in \mathcal{V}$, $v \in \mathcal{R}(V)$. We denote a \textbf{causal formula} as $\varphi^{\mathcal{Y} \leftarrow \textbf{Y}}$ where $\mathcal{Y} \subseteq \mathcal{V}$ and $\varphi$ is a Boolean formula of primitive events. This says that if the variables in $\mathcal{Y}$ were set to values $\textbf{Y}$ (i.e. by \textbf{intervention}) then $\varphi$ would hold. For such a causal formula $\varphi^{\mathcal{Y} \leftarrow \textbf{Y}}$ we write $(M, \textbf{X}) \models \varphi^{\mathcal{Y} \leftarrow \textbf{Y}}$ if $\varphi^{\mathcal{Y} \leftarrow \textbf{Y}}$ is satisfied in causal setting $(M, \textbf{X})$.
\end{definition}

An agent's epistemic state is given by $(\Pr, \mathcal{K}, \mathrm{U})$ where $\mathcal{K}$ is a set of causal settings, $\Pr : \mathcal{K} \rightarrow [0,1]$ is a probability distribution over this set, and $\mathrm{U} : \mathcal{R}(\mathcal{V}) \rightarrow \mathbb{R}_{\geq 0}$ is a utility function.

\theoremstyle{definition}
\begin{definition}
	We define \textbf{how much more likely it is that $\varphi$ will result from performing an action $a$ than from action $a'$} using:
	$$
	    \delta_{a,a',\varphi} = 
	    \max\Bigg(\bigg[  
	    \sum_{(M, \textbf{X}) \models \varphi^{A \leftarrow a}} \Pr(M, \textbf{X})
	    -  
	    \sum_{(M, \textbf{X}) \models \varphi^{A \leftarrow a'}} \Pr(M, \textbf{X})
	    \bigg],0\Bigg)
	$$
	where $A \in \mathcal{V}$ is a variable identified in order to capture an action of the agent.
\end{definition}

The costs of actions are measured with respect to a set of outcome variables $\mathcal{O} \subseteq \mathcal{V}$ whose values are determined by an assignment to all other variables. $\textbf{O}_{(M, \textbf{X})}^{A \leftarrow a}$ denotes the setting of the outcome variables  when action $a$ is performed in causal setting $(M, \textbf{X})$ and $\textbf{V}_{(M, \textbf{X})}^{A \leftarrow a}$ denotes the corresponding setting of the endogenous variables more generally.

\theoremstyle{definition}
\begin{definition}
	The (expected) \textbf{cost of $a$ relative to $\mathcal{O}$} is:
	$$c(a) = 
	\sum_{(M, \textbf{X}) \in \mathcal{K}} 
	\Pr(M, \textbf{X})
	\big[\mathrm{U}(\textbf{V}_{(M, \textbf{X})})
	- 
	\mathrm{U}(\textbf{V}_{(M, \textbf{X})}^{A \leftarrow a})\big]$$
\end{definition}

Finally, HK introduce one last quantity $N$ to measure how important the costs of actions are when attributing blame (this varies according to the scenario). Specifically, as $N \rightarrow \infty$ then $db_{N}(a, a', \varphi) \rightarrow \delta_{a,a',\varphi}$ and thus the less we care about cost. Note that blame is assumed to be non-negative and so it is required that $N > \max_{a \in A}c(a)$.

\theoremstyle{definition}
\begin{definition}
	The \textbf{degree of blameworthiness of $a$ for $\varphi$ relative to $a'$} (given $c$ and $N$) is:
	$$db_{N}(a, a', \varphi) = 
	\delta_{a,a',\varphi}
	\frac{N - \max(c(a') - c(a), 0)}{N}$$
	The overall \textbf{degree of blameworthiness of $a$ for $\varphi$} is then:
	$$db_{N}(a, \varphi) = 
	\max_{a' \in \mathcal{R}(A) \setminus \{a\} } 
	db_{N}(a, a', \varphi)$$
\end{definition}

\subsection{PSDDs}
\label{psdds}

Since, in general, probabilistic inference is intractable \cite{DBLP:journals/jair/BacchusDP09}, tractable learning has emerged as a recent paradigm where one attempts to learn classes of Arithmetic Circuits (ACs), for which exact inference is tractable \cite{gens2013learning,kisa2014probabilistic}.\footnote{It is important to note that this learning framework itself is \textit{approximate}, based on log-likelihoods,  and that tractability guarantees are not always extended to the exact learning of ACs \cite{DBLP:conf/colt/Volkovich16}.} In particular, we use Probabilistic Sentential Decision Diagrams (PSDDs) \cite{kisa2014probabilistic} which are tractable representations of a probability distribution over a propositional logic theory (a set of sentences in propositional logic) represented by a Sentential Decision Diagram (SDD). SDDs are in turn based on vtrees \cite{darwiche2011sdd}. PSDDs thus represent a complete, canonical class with respect to distributional representation, but can also be naturally learnt with the inclusion of logical constraints or background knowledge.\footnote{In this work we refer to PSDDs as \textit{statistical relational models} as we learn them in the presence of logical constraints (encoding relations), but in the absence of such constraints it is more correct to call them purely \textit{statistical models}.} We now provide a brief, formal overview of SDDs and PSDDs and subsequently include a small example in Subsection \ref{example} in order to better illustrate their syntax and semantics. Relationships to other tractable probabilistic models within statistical relational learning are discussed in Section \ref{related}.

\theoremstyle{definition}
\begin{definition}
	A \textbf{vtree} $V$ for a set of variables $\mathcal{X}$ is a full binary tree  whose leaves are in a one-to-one correspondence with the variables in $\mathcal{X}$.
\end{definition}

\theoremstyle{definition}
\begin{definition}
	$S$ is an \textbf{SDD} that is normalised for a vtree $V$ over variables $\mathcal{X}$ if and only if one of the following holds:
	\begin{itemize}
		\item $S$ is a \textbf{terminal node} such that $S = \top$ or $S = \bot$.
		\item $S$ is a terminal node such that $S = X$ or $S = \neg X$ and $V$ is a leaf node corresponding to variable $X$.
		\item $S$ is a \textbf{decision node} $(p_1,s_1),...,(p_k,s_k)$ where \textbf{primes} $p_1,...,p_k$ are SDDs corresponding to the left sub-vtree of $V$, \textbf{subs} $s_1,...,s_k$ are SDDs corresponding to the right sub-vtree of $V$, and $p_1,...,p_k$ form a \textbf{partition} (the primes are consistent, mutually exclusive, and their disjunction $p_1 \lor...\lor p_k$ is valid).
	\end{itemize}
	We refer to each $(p_i,s_i)$ as an \textbf{element} of a decision node. Each terminal node corresponds to its literal or truth symbol and each decision node $(p_1,s_1),...,(p_k,s_k)$ corresponds to the sentence $\bigvee^{k}_{i = 1} (p_i \land s_i)$. $S$ represents a theory (which can be viewed as a set of logical constraints) in that the root of $S$ evaluates to true if and only if the assignment of values to the variables in $\mathcal{X}$ are consistent with that theory.
\end{definition}

Note that in an SDD (and therefore in a PSDD), for any possible assignment of values \textbf{X} to the variables $\mathcal{X}$ that the SDD ranges over, at each decision node  $(p_1,s_1),..., (p_k,s_k)$ at most one prime $p_i$ evaluates to true. In fact, though not strictly necessary, we also make the simplifying assumption that at least one (and thus exactly one) prime evaluates to true for any possible assignment. For such an assignment \textbf{X} we write $\textbf{X} \models p_i$. Further, for any decision node corresponding to a node $v$ in the vtree, the variables $\mathcal{X}_l$ under the left sub-vtree and the variables $\mathcal{X}_r$ under the right sub-vtree partition the set of variables $\mathcal{X}$ in the vtree rooted at $v$, and hence the primes $p_1,...,p_k$ are sentences over $\mathcal{X}_l$ and the subs $s_1,...,s_k$ are sentences over $\mathcal{X}_r$.

\theoremstyle{definition}
\begin{definition}
	A \textbf{PSDD} $P$ is a normalised SDD $S$ (for some vtree $V$) with the following parameters:
	\begin{itemize}
		\item For each decision node $(p_1,s_1),...,(p_k,s_k)$ and prime $p_i$ a non-negative parameter $\theta_i$ such that $\sum^{k}_{i = 1} \theta_i = 1$ and $\theta_i = 0$ if and only if $s_i = \bot$.
		\item For each terminal node $\top$ a parameter $\theta$ such that $0 < \theta < 1$ (denoted as $X:\theta$ where $X$ is the variable of the vtree node that $\top$ is normalised for).
	\end{itemize}
	These parameters then describe the probability distribution over the SDD theory as follows. For each node $n$ in $P$, normalised for some vtree node $v$ in $V$, we have a distribution $\Pr_n$ over the set of variables $\mathcal{X}$ in the vtree rooted at $v$ where:
	\begin{itemize}
	    \vspace{10pt}
		\item If $n$ is a terminal node and $v$ has variable $X$:
        \hspace{7pt} \begin{tabular}{l l l}
            \toprule
			$n$ & ${\Pr}_n(X)$ & ${\Pr}_n(\neg X)$\\
			\midrule
			$X$ & $1$ & $0$\\
			$\neg X$ & $0$ & $1$\\
			$X:\theta$ & $\theta$ & $1 - \theta$\\
			$\bot$ & $0$ & $0$\\
			\bottomrule
		\end{tabular}
		\vspace{10pt}
		\item If $n$ is a decision node $(p_1,s_1),...,(p_k,s_k)$ with parameters $\theta_1 ,..., \theta_k$ and $v$ has variables $\mathcal{X}_l$ in its left sub-vtree and variables $\mathcal{X}_r$ in its right sub-vtree: 
        $${\Pr}_n(\textbf{X}_l,\textbf{X}_r)
		= \sum_{i=1}^k \theta_i {\Pr}_{p_i}(\textbf{X}_l){\Pr}_{s_i}(\textbf{X}_r)
		= \theta_j {\Pr}_{p_j}(\textbf{X}_l){\Pr}_{s_j}(\textbf{X}_r)$$
		for the single $j$ such that $\textbf{X}_l \models p_j$.
	\end{itemize}
	\label{psdddef}
\end{definition}

Most significantly, probabilistic queries, such as conditionals and  marginals, can be computed in time linear in the size of the model. PSDDs can also be learnt from data \cite{liang2017learning}, possibly with the inclusion of  logical constraints standing for  background knowledge. The ability to encode logical constraints into the model (unlike in other tractable probabilistic models, such as the more common Sum-Product Network \cite{poon2011sum}, for example) directly enforces sparsity which in turn can lead to increased accuracy and decreased size. A small selection of ethical considerations relating to the possible use of constraints within our learning framework are discussed in Section \ref{discussion}. Aside from this, the intuitive interpretation of both local and global semantics that can be given to the parameters in a PSDD allows for a degree of explainability not found in other deep probabilistic architectures \cite{kisa2014probabilistic}. A final advantage of PSDDs with respect to our work is that their underlying logical representation makes them particularly conducive to our embedding of the structural equations framework (though existing work in this area is still in its early stages \cite{Papantonis2019}), as we explain further in Section \ref{varz}.

\subsection{Example}
\label{example}

Here we provide a simple worked example demonstrating each of the two frameworks above (we refer the reader to the original works for more extensive examples \cite{kisa2014probabilistic,halpern2018towards}), though this subsection may safely be skipped with respect to our results and later discussion. The experimental results in Section \ref{experiments} provide examples of our particular embedding of HK's framework, and more realistic applications are discussed in Section \ref{discussion}. 

Consider the following decision-making scenario, with four binary variables, in which Alfred is walking to work and is not sure if it will rain ($R$); he thinks the probability that it will is 0.5. If he decides to go back and collect his umbrella ($U$) there is a probability (again, 0.5) he will be late ($L$). However, more important than his being on time is whether he arrives at work wet ($W$) or dry ($\neg W$). In HK's framework we have $\mathcal{X} = \{R\}$ and $\mathcal{V} = \{U, W, L\}$. Let $M_1$ contain these variables and the structural equations $\mathcal{F}_1$ such that Alfred is late to work due to his going back, and $M_2$ include structural equations $\mathcal{F}_2$ such that he is not late despite going back.\footnote{Here we assume that the `default' option in any one specific causal model is to collect the umbrella if it will rain and not when it won't, though of of course Alfred does not have this perfect information.} Then $\mathcal{K} = \{(M_1, \neg R), (M_1, R), (M_2, \neg R), (M_2, R)\}$ and $\Pr$ is such that $\Pr(M, \textbf{X}) = 0.25$ for all $(M, \textbf{X}) \in \mathcal{K}$. We define our utility function such that being on time gives utility 2 and remaining dry gives utility 3 (meaning overall utilities fall in the range $[0,5]$). A causal graph and set of structural equations for $M_1$ and the context $R = 1$ is given in Figure \ref{fig:umbrellagraph}.

\begin{figure}[h]
    \centering
    \begin{minipage}[c]{0.3\linewidth}
        \vspace{0pt}
        \centering
        \begin{tikzpicture}[
        node distance =2cm and 2cm ,
            on grid ,
        every node/.style={
            draw, circle, minimum size=0.5cm, minimum width=0.75cm, minimum height=0.75cm, inner sep=0.5mm}]
        \node (R) {$R$};
        \node (L) [right of = R] {$L$};
        \node (W) [below of = R] {$W$};
        \node (U) [right of = W] {$U$};
        \path (R) edge [->] (W);
        \path (U) edge [->] (W);
        \path (R) edge [->] (U);
        \path (U) edge [->] (L);
        \end{tikzpicture}
    \end{minipage}
    \begin{minipage}[c]{0.65\linewidth}
        \vspace{0pt}
        \centering
        \begin{tabular}{l l l}
            \toprule
            \textbf{Variable} & \textbf{Equation} & \textbf{Note}\\
    		\midrule
    		$R$ & 1 & Due to the context $R = 1$\\
    		$U$ & $R$ & The `default' strategy\\
    		$L$ & $U$ & Varies between $M_1$ and $M_2$\\
    		$W$ & $R(1 - U)$ & The same in $M_1$ and $M_2$\\
        	\bottomrule
        \end{tabular}
    \end{minipage}
    \caption{A causal graph and set of structural equations representing the causal setting $(M_1, R)$ in which it rains and Alfred is late to work if he goes back to collect his umbrella.}
    \label{fig:umbrellagraph}
\end{figure}

Suppose we wish to compute $db_{N}(U, \neg U, L = 1)$ with $N=2$, say: how blameworthy Alfred is for being late to work because he went back to get his umbrella. Note that $\{(M,\textbf{X}) : (M,\textbf{X}) \models (L=1)^{U \leftarrow 1}\} = \{(M_1, \neg R), (M_1, R)\}$ and $\{(M,\textbf{X}) : (M,\textbf{X}) \models (L=1)^{U \leftarrow 0}\} = \varnothing$. Thus we have $\delta_{U,\neg U,L=1} = \max\big([(0.25 + 0.25) - 0], 0\big) = 0.5$. Next, we sum over the differences in expected utility across all causal models to find that $c(U) = 0.25[5 - 3] + 0.25[3 - 3] + 0.25[5 - 5] + 0.25[5 - 5] = 0.5$. Similarly, we have $c(\neg U) = 1$. Substituting these values into the final equation gives $db_{2}(U, \neg U, L = 1) = \delta_{U,\neg U,L=1}\frac{2 - \max(c(\neg U) - c(U), 0)}{2} = 0.375$.

\begin{figure}[h]
\centering
\includegraphics[width=0.95\textwidth]{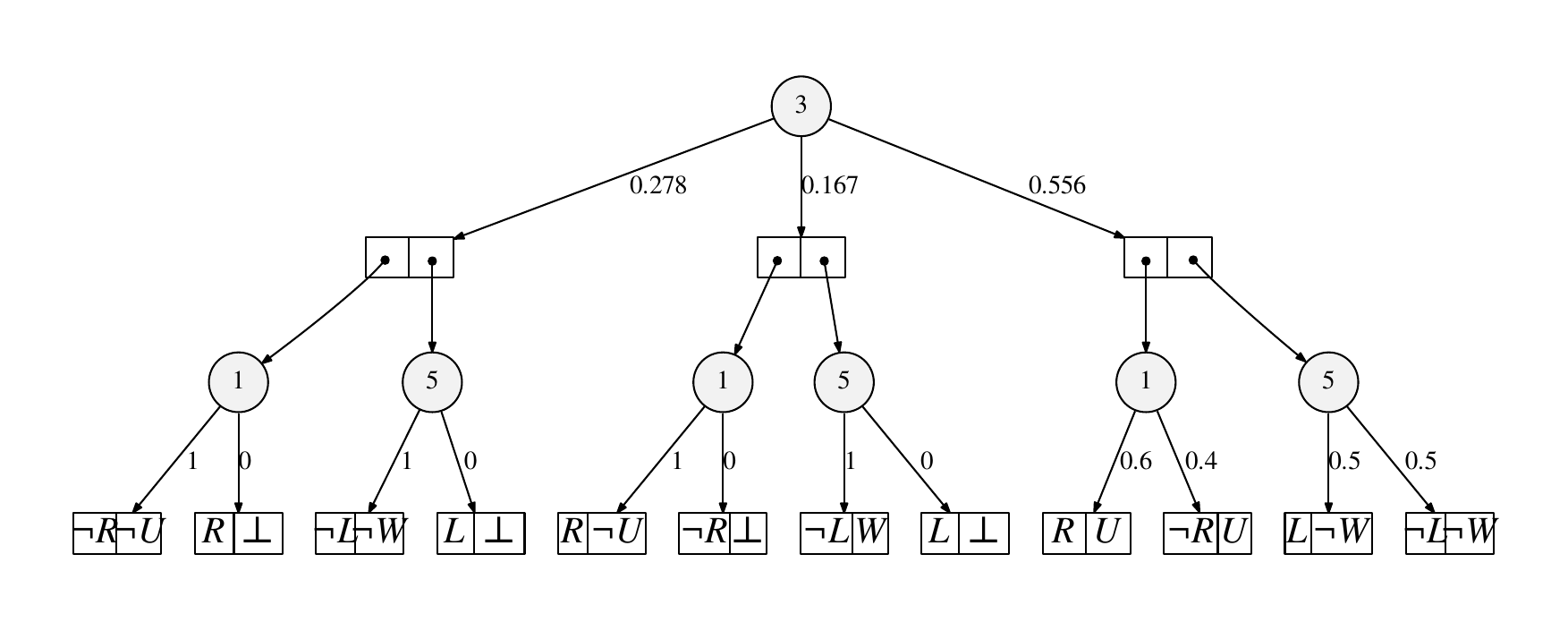}
\caption{A PSDD encoding all features of the example data and constraints as defined above.}
\label{fig:examplepsdd}
\end{figure}

Now, using the same variables and probabilities as above, let us imagine we have some dataset of decision-making scenarios (gathered from irrational agents, if we are to assume the same utility function as above) in which the umbrella is collected with probability 0.667 when it rains and probability 0.444 otherwise. We might also wish to constrain our model with background knowledge  such that the decision-maker arrives to work wet if and only if it is raining and they don't have their umbrella ($W \leftrightarrow (R \land \neg U)$, akin to the structural equation $W = R(1 - U)$, for example) and that they cannot be late if they don't go back for their umbrella ($\neg U \rightarrow \neg L$, consistent with structural equations $L=U$ in $M_1$ and $L=0$ in $M_2$). Combining these data and constraints allows our system to learn the small PSDD shown in Figure \ref{fig:examplepsdd} (note that the model would typically be further compacted by removing superfluous branches or nodes and possibly joining some of the remaining nodes, but for ease of presentation we have not done so here).

\section{Blameworthiness Via PSDDs}
\label{embedding}

We aim to leverage the learning of PSDDs, their tractable query interface, and their ability to handle domain constraints for inducing models of moral scenarios.\footnote{Our technical development can leverage both parameter and (possibly partial) structure learning for PSDDs. Of course, learning causal models is a challenging problem \cite{DBLP:journals/corr/abs-1805-09697}, and in this regard, probabilistic structure learning is not assumed to be a recipe for causal discovery in general \cite{pearl1998graphical}. Rather, we learn a probabilistic model from data without making any claims about the causal dependencies between variables that may be induced and show how, under certain assumptions discussed later, we are able to perform causal reasoning such as conditioning on intervention.} This is made possible by means of an embedding that we sketch below in terms of the key components of our models and computations: variables, probabilities, utilities, and finally costs and blameworthiness. In each subsection we also discuss any assumptions required and choices made. At the outset, we reiterate that we do not introduce new definitions here, but show how an existing one, that of HK, can be embedded within a learning framework. Where there is any chance of ambiguity we denote the original definitions with a superscript $^{HK}$.

\subsection{Variables}
\label{varz}

We first distinguish between scenarios in which we do and do not model \textit{outcome variables}. In both cases we have exogenous variables $\mathcal{X}$, but in the former the endogenous variables $\mathcal{V}$ are partitioned into decision variables $\mathcal{D}$ and outcome variables $\mathcal{O}$, and in the latter we have $\mathcal{V} = \mathcal{D} = \mathcal{O}$ (this does not affect the notation in our later definitions, however). This is because we do not assume that outcomes can always be recorded, and in some scenarios it makes sense to think of decisions as an end in themselves.

Our \textit{range function} $\mathcal{R}$ is defined by the scenario we model, but in practice we one-hot encode the variables and so the range of each is simply $\{0,1\}$. A subset (possibly empty) of the \textit{structural equations} in $\mathcal{F}$ is implicitly encoded within the structure of the SDD underlying the PSDD, corresponding to the logical constraints that remain true in every causal model $M$. The remaining equations are those that vary depending on the causal model. Each possible assignment $\textbf{V}$ (in other words $\textbf{D}$ and $\textbf{O}$) given $\textbf{X}$ corresponds to a set of structural equations that combine with those encoded by the SDD to determine the values of the variables in $\mathcal{V}$ given $\textbf{X}$, as we make the trivial assumption that all parentless variables are considered exogenous. The PSDD then corresponds to the probability distribution $\Pr$ over $\mathcal{K}$, compacting everything neatly into a single structure, as described in Subsection \ref{probz}.

To be more precise regarding our use of the word `implicitly' in the above paragraph, structural equations can be viewed as encoding (in a specific functional form) dependencies and independencies between variables. Such dependencies are similarly (though not identically) captured by propositional formulae involving multiple variables. In fact, just as one may read off independencies from  casual graphs representing sets of structural equations using the well-known \emph{d-separation} criterion \cite{pearl2009causal}, it is also possible to read off independencies from the structure of a PSDD (though this feature is not necessary for our purposes). Such structures, more generally,  enable \emph{context-specific independence} \cite{Boutilier1996,kisa2014probabilistic}. We further note that the framework of HK essentially involves events of the form $V = v$ (see Definition \ref{prim}) which can be viewed instead simply as propositions \cite{halpern2005causes} (especially when considering binary variables, as in our equivalent one-hot encoding), in turn meaning that structural equations represent logical formulae. While this suits our choice of PSDDs well, we remark that in general structural equations may be far more complex and therefore less amenable to the embedding we describe here.

Returning to our example in Figure \ref{fig:umbrellagraph}, if we were to enforce the constraint $U \leftrightarrow R$ in the PSDD in Figure \ref{fig:examplepsdd} then this would capture the dependency $U = R$ or $R = U$ (where `=' is directional in the standard sense of structural equations \cite{pearl2009causal}).\footnote{We have offered a very simple instance here, as in the general case logical constraints in a PSDD serve to \emph{rule out} sets of structural equations (if they are logically inconsistent with the constraints) rather than capture them directly (and so we would have to provide a set of possible equations as an example for a single logical constraint). The link between structural equations and constraints is thus by no means `one-to-one'.} We note, however, that for the purposes of learning a distribution from data, the direction in this structural equation does not matter per se. Where the direction is revealed, and is critically important, is when intervening on variables. For example, intervening on $R$ would change the value of $U$ when the equation is $U = R$ but not when $R = U$. This difference underlies why in general it is not possible to answer arbitrary causal queries using a probabilistic model. In this work, however, our queries are of a specific form which means that a probabilistic model is sufficient for our purposes (as explained in the following section). Therefore, although our models do not encode directionality in the same way as structural equations, they are nonetheless suitable for our embedding. On a related theme, we also note that the opposite direction of obtaining structural equations from PSDDs is also non-trivial, though there is recent work in this direction \cite{Papantonis2019}.

Our critical assumption here is that the \textit{signature} $\mathcal{S} = (\mathcal{X, V, R})$ (the variables and the values they may take) remains the same in all models, although the structural equations $\mathcal{F}$ (the ways in which said variables are related) may vary. Given that each model represents an agent's uncertain view of a decision-making scenario we do not think it too restrictive to keep the elements of this scenario the same across the potential eventualities, so long as the way these elements interact may differ. Indeed, learning PSDDs from decision-making data requires that the data points measure the same variables each time.

\subsection{Probabilities}
\label{probz}

Thus, by partially encoding the possible sets of structural equations governing the variables in the domain (those not ruled out by the logical constraints on the PSDD), the represented \textit{distribution} $\Pr: \mathcal{R}(\mathcal{X} \cup \mathcal{D} \cup \mathcal{O}) \rightarrow [0,1]$ ranges over assignments to variables instead of a set of causal models $\mathcal{K}$. As a slight abuse of notation we write $\Pr(\textbf{X},\textbf{D},\textbf{O})$. The key observation needed to translate between these two distributions (we denote the original as $\Pr^{HK}$) is that each set of structural equations $\mathcal{F}$ together with a context $\textbf{X}$ deterministically leads to a unique, complete assignment $\textbf{V}$ of the endogenous variables, which we write (abusing notation slightly again) as $(\mathcal{F},\textbf{X}) \models \textbf{V}$. In general there may be many such sets of equations that lead to the same assignment (in other words, many possible sets of rules governing the world which, given a context, produce the same result), which we may write as $\{\mathcal{F} : (\mathcal{F},\textbf{X}) \models \textbf{V}\}$. This observation relies on our assumption above, which implies that for any causal model $(M, \textbf{X})$ we in fact have $\Pr^{HK}(M, \textbf{X}) = \Pr^{HK}((\mathcal{S},\mathcal{F}), \textbf{X}) = \Pr^{HK}(\mathcal{F}, \textbf{X})$, as the signature $\mathcal{S}$ is the same in all models. Hence, for any context $\textbf{X}$ and any (possibly empty) assignment $\textbf{Y}$ for $\mathcal{Y} \subseteq \mathcal{V}$ we may translate between the two distributions as follows:
	$$
	        \Pr(\textbf{X},\textbf{Y}) 
	        = \sum_{(\mathcal{F},\textbf{X}) \models \textbf{Y}} {\Pr}^{HK}(\mathcal{F},\textbf{X})
	$$
Given our assumptions and observations described above, the following proposition is immediate.

\begin{proposition}
\label{probprop}
Let ${\Pr}^{HK}$ be a probability distribution over a set of causal settings $\mathcal{K}$. Further, assume that the signature $\mathcal{S} = (\mathcal{X, V, R})$ in each causal setting $M = (\mathcal{S}, \mathcal{F})$ remains fixed. Then there exists a PSDD $P$ representing a distribution $\Pr$ over the variables in $\mathcal{X}$ and $\mathcal{V}$ such that for any context $\textbf{X}$, the joint probability of $\textbf{Y}$ also occurring (where $\mathcal{Y} \subseteq \mathcal{V}$) is the same under both ${\Pr}^{HK}$ and $\Pr$.
\end{proposition}

We view a Boolean formula of primitive events (possibly resulting from decision $A$) as a function $\varphi: \mathcal{R}(\mathcal{Y}) \rightarrow \{0,1\}$ that returns 1 if the original formula over $\mathcal{Y} \subseteq \mathcal{V}$ is satisfied by the assignment, or 0 otherwise. Here, the probability of $\varphi$ occurring given that action $a$ is performed (i.e. conditioning on \textit{intervention})  $\sum_{(M, \textbf{X}) \models \varphi^{A \leftarrow a}} \Pr^{HK}(M, \textbf{X})$ given by HK can be written more simply as $\Pr(\varphi~\vert~do(a))$.  Note that in general, it is not the case that $\Pr(\varphi~\vert~do(a)) = \Pr(\varphi~\vert~a)$, where $\Pr(\varphi~\vert~a)$ is defined as the standard conditioning on \textit{observation}. However, given the specific nature of the causal models in our framework which capture sequential/temporal moral decision-making scenarios (in which one or more decisions are made in some context, each producing one or more outcomes), and given that the quantities we calculate only require us to intervene on decision variables, we are able to make use of a certain technical result and compute intervention conditionals in terms of observation conditionals. We note here that our mild assumptions below on the structure of the domain refers to the structure of \emph{the actual data-generating process} rather than referring to some feature of the learnt PSDD.

\begin{figure}[h]
    \centering
    \begin{tikzpicture}[
        node distance =2.5cm and 3cm ,
            on grid ,
        every node/.style={
            draw, minimum size=0.5cm, inner sep=0.5mm}]
        \node (X) [red, ellipse, dashed, minimum height=0.75cm, minimum width=6cm] {$\mathcal{X}$};
        \node (D) [circle, minimum width=0.75cm, minimum width=0.75cm, below of = X] {$A$};
        \node (D_pre) [red, ellipse, dashed, minimum height=0.75cm, minimum width=2cm, left of = D] {$\mathcal{D}_{pre}$};
        \node (D_post) [blue, ellipse, dashed, minimum height=0.75cm, minimum width=2cm, right of = D] {$\mathcal{D}_{post}$};
        \node (O_pre) [red, ellipse, dashed, minimum height=0.75cm, minimum width=3cm,  below of = D_pre] {$\mathcal{O}_{pre}$};
        \node (O_post) [blue, ellipse, dashed, minimum height=0.75cm, minimum width=3cm,  below of = D_post] {$\mathcal{O}_{post}$};
        \path (X) edge [->] (D_pre);
        \path (X) edge [->] (D);
        \path (X) edge [->] (D_post);
        \path (D_pre) edge [->] (D);
        \path (D_pre) edge [bend left = 30, ->] (D_post);
        \path (D) edge [->] (D_post);
        \path (D) edge [->] (O_post);
        \path (D_pre) edge [->, bend right = 10] (O_pre);
        \path (O_pre) edge [->, bend right = 10] (D_pre);
        \path (D_pre) edge [->] (O_post);
        \path (D_post) edge [->, bend right = 10] (O_post);
        \path (O_post) edge [->, bend right = 10] (D_post);
        \path (X) edge [->, bend right = 60] (O_pre);
        \path (X) edge [->, bend left = 60] (O_post);
        \path (O_pre) edge [->] (D_post);
        \path (O_pre) edge [->] (D);
        \path (O_pre) edge [->] (O_post);
    \end{tikzpicture}
    \caption{A causal graph representing the form of sequential moral decision-making scenario we consider in this work. Dashed edges indicate sets of variables (which may also contain other arrows between nodes; it is assumed that such arrows break the apparent cyclicity between $\mathcal{D}_{pre}$ and $\mathcal{D}_{post}$, and $\mathcal{O}_{pre}$ and $\mathcal{O}_{post}$) and solid edges indicate single variables. Arrows represent causal connections. The set of variables $\mathcal{PRE}$ is highlighted in red and blocks all back-door paths between $\mathcal{POST}$ (highlighted in blue) and the decision variable in question, $A$, thus forming a sufficient set.}
    \label{fig:causalgraph}
\end{figure}

To see this, note that in the causal graph of such a decision-making scenario (see Figure \ref{fig:causalgraph}), the ancestors (by which we mean those nodes with a directed causal path to the node in question) of a decision variable $A$, representing some action, are a (possibly non-proper) subset of the context variables $\mathcal{X}$, any preceding decision variables $\mathcal{D}_{pre}$, and any outcome variables that have been determined $\mathcal{O}_{pre}$, where we write $\mathcal{PRE} = \mathcal{X}\cup\mathcal{D}_{pre}\cup\mathcal{O}_{pre}$ to denote this set. Note also that any remaining decisions $\mathcal{D}_{post}$ and outcomes $\mathcal{O}_{post}$, where we similarly write $\mathcal{POST} = \mathcal{D}_{post}\cup\mathcal{O}_{post}$, are in turn caused by the variables in $\mathcal{PRE}\cup\{D\}$. This is true simply in virtue of the form of decision-making scenarios that we consider in this work. Given this, we may use the \textit{back-door criterion} \cite{pearl2009causal} with $\mathcal{PRE}$ as a \textit{sufficient set} (meaning that no element of $\mathcal{PRE}$ is a descendant of $A$ and that $\mathcal{PRE}$ blocks all back-door paths from $\mathcal{POST}$ to $A$) to write:
$$\Pr(\textbf{POST}~\vert~do(a)) = \sum_{\textbf{PRE}} \Pr(\textbf{POST}~\vert~a, \textbf{PRE})\Pr(\textbf{PRE})$$
Here we write \textbf{PRE} and \textbf{POST} for instantiations of $\mathcal{PRE}$ and $\mathcal{POST}$ respectively, just as for a variable set $\mathcal{Y}$ and an instantiation \textbf{Y}. Note that $\mathcal{PRE}$, $\mathcal{POST}$, and $\{A\}$ partition the full set of variables, and in the case where there is only a single decision, $\mathcal{D}=\{A\}$, then we have simply $\mathcal{PRE} = \mathcal{X}$ and $\mathcal{POST} = \mathcal{O}$. Given the equality above we may thus compute the quantity $\Pr(\varphi~\vert~do(a))$ as follows:
\begin{equation*}
    \begin{aligned}
        \Pr(\varphi~\vert~do(a)) 
        &= \sum_{\textbf{POST}}\Pr(\varphi(\textbf{POST})~\vert~do(a)) \\
        &= \sum_{\textbf{POST}} \varphi(\textbf{POST})\Pr(\textbf{POST}~\vert~do(a)) \\
        &= \sum_{\textbf{POST}}\sum_{\textbf{PRE}} \varphi(\textbf{POST})\Pr(\textbf{POST}~\vert~a, \textbf{PRE})\Pr(\textbf{PRE})
    \end{aligned}
\end{equation*}
where we again use our mapping between $\Pr$ and $\Pr^{HK}$ given above. With this equivalence we define our term $ \delta_{a,a',\varphi} = 
	    \max([\Pr(\varphi~\vert~do(a)) 
	    -  
	    \Pr(\varphi~\vert~do(a')) 
	    ],0)$ as in HK. In some cases we may wish to calculate blameworthiness in scenarios in which the distribution over contexts is not the same as in our training data. Fortunately, due to our factorised sum above this is as simple as allowing the user of our system the option of specifying a current, alternative distribution over contexts and existing observations $\Pr'(\textbf{PRE})$, which then replaces the term $\Pr(\textbf{PRE})$ in each summand.
	    
We remark here that although the causal structure illustrated in Figure \ref{fig:causalgraph} admits a wide range of sequential moral decision-making scenarios and allows us to compute all of the quantities we need, it is also the case that additional variables and dependencies could invalidate our use of the back-door criterion, and that it is certainly not possible to compute the effects of arbitrary interventions in this model. Returning to our previous example, suppose (as illustrated in the left half of Figure \ref{fig:umbrellaces}) that both Alfred's decision to take his umbrella and whether or not he is late to work depends on whether he sees the bus approaching from out of the window ($B$). If we were, for some reason, unable to condition on $B$ as part of a sufficient set, then the back-door path $L \longleftarrow B \longrightarrow U$ would not be blocked and thus $\Pr(l,w ~\vert~ do(u)) \neq \Pr(l,w ~\vert~ u, r)\Pr(r)$. Similarly if we wanted to, say, condition on an intervention on a non-decision variable then we would not be able to. Consider the slightly modified version of the original scenario in which $L$ also depends on $W$ (perhaps because Alfred cycles to work and is slower when his clothes are wet) in the right half of Figure \ref{fig:umbrellaces}, and consider an intervention $do(w)$. Then $\Pr(l ~\vert~ do(w)) \neq \Pr(l ~\vert~ w, r)\Pr(r)$ because there is an unblocked back-door path $L \longleftarrow U \longrightarrow W$.

\begin{figure}[h]
    \centering
    \begin{minipage}[c]{0.45\linewidth}
        \vspace{0pt}
        \centering
        \begin{tikzpicture}[
        node distance =2cm and 2cm ,
            on grid ,
        every node/.style={
            draw, circle, minimum size=0.5cm, minimum width=0.75cm, minimum height=0.75cm, inner sep=0.5mm}]
        \node (R) [red] {$R$};
        \node (L) [blue, right of = R] {$L$};
        \node (W) [blue, below of = R] {$W$};
        \node (U) [right of = W] {$U$};
        \node (B) [right of = U] {$B$};
        \path (R) edge [->] (W);
        \path (U) edge [->] (W);
        \path (R) edge [->] (U);
        \path (U) edge [->] (L);
        \path (B) edge [->] (U);
        \path (B) edge [->] (L);
        \end{tikzpicture}
    \end{minipage}
    \begin{minipage}[c]{0.45\linewidth}
        \vspace{0pt}
        \centering
        \begin{tikzpicture}[
        node distance =2cm and 2cm ,
            on grid ,
        every node/.style={
            draw, circle, minimum size=0.5cm, minimum width=0.75cm, minimum height=0.75cm, inner sep=0.5mm}]
        \node (R) [red] {$R$};
        \node (L) [blue, right of = R] {$L$};
        \node (W) [below of = R] {$W$};
        \node (U) [right of = W] {$U$};
        \path (R) edge [->] (W);
        \path (U) edge [->] (W);
        \path (R) edge [->] (U);
        \path (U) edge [->] (L);
        \path (W) edge [->] (L);
        \end{tikzpicture}
    \end{minipage}
    \caption{Two causal graphs indicating how the back-door criterion can be violated when extra variables are added that we are unable to condition on (left) and when the intervention in question is not made on a decision node (right).}
    \label{fig:umbrellaces}
\end{figure}

\subsection{Utilities}

We now consider the \textit{utility function} $\mathrm{U}$, the output of which we assume is normalised to the range $[0,1]$.\footnote{This has no effect on our calculations as we only use cardinal utility functions with bounded ranges, which are invariant to positive affine transformation.} For simplicity we (trivially) restrict our utility functions to be over outcomes $\textbf{O} = (O_1,...,O_n)$ (and optionally parameterised using contexts $\textbf{X}$) instead of the full set of endogenous variables. In our implementation we allow the user to input an existing utility function or to learn one from data. In the latter case the user further specifies whether or not the function should be context-relative, i.e. whether we have $\mathrm{U}(\textbf{O})$ or $\mathrm{U}(\textbf{O};\textbf{X})$ (our notation) as, in some cases, how good a certain outcome $\textbf{O}$ is naturally depends on the context $\textbf{X}$. Similarly, the user also decides whether the function should be linear in the outcome variables, in which case the final utility is $\mathrm{U}(\textbf{O}) = \sum_{i}\mathrm{U}_i(O_i)$ or $\mathrm{U}(\textbf{O};\textbf{X}) = \sum_{i}\mathrm{U}_i(O_i;\textbf{X})$ respectively (where we assume that each $\mathrm{U}_i(O_i;\textbf{X})$ or $\mathrm{U}_i(O_i)$ is non-negative). Here the utility function is simply a vector of weights and the total utility of an outcome is the dot product of this vector with the vector of outcome variables $(O_1,...,O_n)$.

When learning utility functions, the key assumption we make (before normalisation) is that \textit{the probability of a certain decision being made given a context is proportional to some function of the expected utility of that decision in the context}, i.e. $\Pr(\textbf{D}~\vert~\textbf{X}) \propto f(\sum_{\textbf{O}}\mathrm{U}(\textbf{O})\Pr(\textbf{O}~\vert~\textbf{D},\textbf{X}))$. Note that here a decision is a general assignment \textbf{D} and not a single action $a$, and $\mathrm{U}(\textbf{O})$ may be context-relative and/or linear in the outcome variables. In our implemented demonstration system we make the simplifying assumption that $f$ is the identity function (and thus the proportionality represents a linear relationship), however this is by no means necessary. In general we may choose any invertible function $f$ (on the range $[0,1]$) and simply apply $f^{-1}$ to each datum $\Pr(\textbf{D}~\vert~\textbf{X})$ before fitting our utility function, the process of said fitting being described in Section \ref{implementation}.\footnote{In general we should expect $f$ to be a positive monotonic transformation with non-negative range so as to preserve the ordinality of utilities.} For example, using $f(x) = \exp(x) - 1$ allows us to capture (a slightly modified version of) the commonly-used \textit{Logistic Quantal Response} model of bounded rationality, sometimes referred to as \textit{Boltzmann Rationality}, in which the likelihood of a certain decision is proportional to the exponential of the resulting expected utility \cite{mckelvey1995quantal}.

This proportionality assumption is critical to the learning procedure in our implementation, however we believe it is in fact relatively uncontroversial, and can be restated as the simple rationality principle that an agent is more likely to choose a decision that leads to a higher expected utility than one that leads to a lower expected utility. If we view decisions as guided by a utility function, then it follows that the decisions should, on average, be consistent with and representative of that utility function. Of course this is not always true (consider the smoker who wishes to quit but cannot due to their addiction), and attempting to learn the preferences of fallible, inconsistent agents such as humans is a particularly interesting and difficult problem. While outside the scope of our current work, we refer the reader to Evans et al. for a recent discussion \cite{evans2016learning}. We also note here that learning moral preferences from data must be done sensitively, at is it is quite possible the data may include biases that we would typically deem unethical. Space precludes us from discussing this important issue further, but it is undoubtedly a key concern in practice for any method that learns from human decision-making.

\subsection{Costs and Blameworthiness}

Finally, we adapt the \textit{cost function} given in HK, denoted here by $c^{HK}$. As actions do not deterministically lead to outcomes in our work, we cannot use $\textbf{O}_{(M,\textbf{X})}^{A \leftarrow a}$ to represent the specific outcome when decision $a$ is made (in some context $(M,\textbf{X})$). For our purposes it suffices to use: 

$$c(a) = - \sum_{\textbf{O}} \mathrm{U}(\textbf{O})\Pr(\textbf{O}~\vert~do(a)) = - \sum_{\textbf{O},\textbf{PRE}} \mathrm{U}(\textbf{O})\Pr(\textbf{O}~\vert~a,\textbf{PRE})\Pr(\textbf{PRE})$$

Again, $\mathrm{U}$ may be context-relative and/or linear in the outcome variables. This is simply the negative expected utility over all contexts and preceding decisions/outcomes, conditioning by intervention on decision $A \leftarrow a$. By assuming as before that action $a$ is causally influenced only by the variables in set $\mathcal{PRE}$ (i.e. $\mathcal{PRE}$ is a \textit{sufficient set} for $A$) we may again use the \textit{back-door criterion} \cite{pearl2009causal} to write $\Pr(\textbf{O}~\vert~do(a)) = \sum_{\textbf{PRE}}\Pr(\textbf{O}~\vert~a,\textbf{PRE})\Pr(\textbf{PRE})$. With this useful translation between conditioning on intervention and conditioning on observation, alongside our earlier result converting between $\Pr^{HK}$ and $\Pr$, it is a straightforward exercise in algebraic manipulation to show the following proposition.

\begin{proposition}
\label{costprop}
Let $c^{HK}$ be a cost function determined using a distribution ${\Pr}^{HK}$ and utility function $U$. Then, given an equivalent distribution $\Pr$ (via the assumptions and result of Proposition \ref{probprop}) and the assumption that $\mathcal{X}$ forms a sufficient set for any action variable $A$, the cost function $c$ determined using $\Pr$ and $U$ is such that for any values $a, a'$ of $A$: $c(a') - c(a) = c^{HK}(a') - c^{HK}(a)$.
\end{proposition}

As $db_{N}(a, a', \varphi) = \delta_{a,a',\varphi} \frac{N - \max(c(a') - c(a), 0)}{N}$ it follows that our cost function is equivalent to the one in HK with respect to determining blameworthiness scores. Again, in our implementation we also give the user the option of updating the distribution over contexts and previously observed variables $\Pr(\textbf{PRE})$ to some other distribution $\Pr'(\textbf{PRE})$ so that the current model can be re-used in different scenarios. Given $\delta_{a,a',\varphi}$, $c(a)$, and $c(a')$, both $db_{N}(a, a', \varphi)$ and $db_{N}(a, \varphi)$ are computed as in HK, although we instead require that $N > -min_{a \in A} c(a)$ (the equivalence of this condition to the one in HK is trivial). With this the embedding is complete.

\begin{proposition}
Let $\Pr$ and $c$ be equivalents of ${\Pr}^{HK}$ and $c^{HK}$ under the assumptions and results described in Propositions \ref{probprop} and \ref{costprop}. Then for any values $a, a'$ of any action variable $A \in \mathcal{D} \subseteq \mathcal{V}$, for any Boolean formula $\varphi$, and any valid measure of cost importance $N$, the values of $\delta_{a,a',\varphi}$, $db_{N}(a, a', \varphi)$, and $db_{N}(a, \varphi)$ are the same in our embedding as in HK.
\end{proposition}

\section{Complexity Results}
\label{complexityresults}

Given our concerns over tractability we provide several computational complexity results for our embedding. Basic results were given in HK, but only in terms of the computations being polynomial in $\vert M \vert$, $\vert\mathcal{K}\vert$, and $\vert\mathcal{R}(A)\vert$ \cite{halpern2018towards}. Here we provide more detailed results that are specific to our embedding and to the properties of PSDDs. The complexity of calculating blameworthiness scores also depends on whether the user specifies an alternative distribution $\Pr'$ in order to consider specific contexts only, although in practice this is unlikely to have a major effect on tractability. Finally, note that here we assume that the PSDD and utility function are given in advance and so we do not consider the computational cost of learning. This parallels the results in HK, in which only the cost of reasoning is considered (there is no mention of how their models are obtained). As mentioned previously, guarantees in the tractable learning paradigm are provided for \textit{tractable inference within learnt models}, but not for the learning procedure itself, which is often approximate \cite{DBLP:conf/colt/Volkovich16}. A summary of our results is given in Table \ref{complexities}.

Here, $O(\vert P \vert )$ is the time taken to evaluate the PSDD $P$ where $\vert P \vert $ is the size of the PSDD, measured as the number of parameters; $O(U)$ is the time taken to evaluate the utility function; and $O(\vert \varphi \vert )$ is the time taken to evaluate the Boolean function $\varphi$, where $\vert \varphi \vert $ measures the number of Boolean connectives in $\varphi$. The proofs of the results above are an easy exercise (we give an informal explanation of each in the following paragraph), though for illustrative purposes we provide one example.

\begin{proposition}
$\delta_{a,a',\varphi}$ can be calculated using our framework with time complexity \newline $O(2^{\vert \mathcal{X} \vert + \vert \mathcal{D} \vert + \vert \mathcal{O} \vert}( \vert \varphi\vert + \vert P \vert ))$.
\end{proposition}

\begin{proof}
First recall that, following the definitions in HK and our embedding from Section \ref{embedding}, in our framework we use:
$$ \delta_{a,a',\varphi} = 
	    \max\Big(\big[\Pr(\varphi~\vert~do(a)) 
	    -  
	    \Pr(\varphi~\vert~do(a')) 
	    \big],0\Big)$$
Where, as was shown in Subsection \ref{probz}: $$ \Pr(\varphi~\vert~do(a)) = \sum_{\textbf{POST}}\sum_{\textbf{PRE}} \varphi(\textbf{POST})\Pr(\textbf{POST}~\vert~a, \textbf{PRE})\Pr(\textbf{PRE})$$
The proof now follows straightforwardly from inspection of the terms involved. Calculating $\Pr(\textbf{PRE})$ and $\Pr(\textbf{POST}~\vert~a, \textbf{PRE})$ can each be done in time $O(\vert P \vert )$, linear in the size of the PSDD $P$ representing the distribution over all variables \cite{kisa2014probabilistic}. $\varphi(\textbf{POST})$ is computed in time $O(\vert \varphi \vert )$, linear in the length of $\varphi$. Thus, forming each summand in the expression for $\Pr(\varphi~\vert~do(a))$ takes time $O(\vert P\vert + \vert \varphi \vert )$, and as each variable being summed over is binary, we need to calculate at most $2^{\vert \mathcal{X} \vert + \vert \mathcal{D}\vert + \vert \mathcal{O}\vert}-1$ such summands, giving us a time complexity for $\Pr(\varphi~\vert~do(a))$ of $O(2^{\vert \mathcal{X} \vert + \vert \mathcal{D} \vert + \vert \mathcal{O} \vert}( \vert \varphi\vert + \vert P \vert ))$. This is the same for our other term $\Pr(\varphi~\vert~do(a'))$, and the remaining arithmetic operations can be performed in constant time, meaning the final complexity of calculating $\delta_{a,a',\varphi}$ is also $O(2^{\vert \mathcal{X} \vert + \vert \mathcal{D} \vert + \vert \mathcal{O} \vert}(\vert\varphi\vert + \vert P \vert ))$.
\end{proof}

\begin{table}[H]
\begin{center}
\begin{tabular}{l l}
\toprule
\textbf{Term} & \textbf{Time Complexity}\\
		\midrule
		$\delta_{a,a',\varphi}$ & $O(2^{\vert \mathcal{X} \vert + \vert \mathcal{D} \vert + \vert \mathcal{O} \vert}( \vert \varphi\vert + \vert P \vert ))$\\
		$c(a)$ & $O(2^{\vert \mathcal{X} \vert + \vert \mathcal{O} \vert}(U+\vert P \vert ))$\\
		$db_{N}(a, a', \varphi)$ & $O(2^{\vert \mathcal{X}\vert + \vert \mathcal{O}\vert}(U + 2^{\vert \mathcal{D} \vert}( \vert \varphi\vert + \vert P \vert )))$\\
		$db_{N}(a, \varphi)$ & $O(\vert \mathcal{R}(A) \vert2^{~\vert~\mathcal{X}\vert + \vert \mathcal{O}\vert}(U + 2^{\vert \mathcal{D} \vert}( \vert \varphi\vert + \vert P \vert )))$\\
		\bottomrule
\end{tabular}
\end{center}
\caption{Time complexities for each of the key terms that we compute. If the user specifies an extra distribution $\Pr'$ over contexts, then the complexity is given by the expressions below with each occurrence of the term $\vert P \vert $ replaced by $\vert P \vert  + Q$, where $O(Q)$ is the time taken to evaluate $\Pr'$.}
\label{complexities}
\end{table}

Note that although we have to evaluate $\varphi$ each time, in practice only a small subset of all possible models will evaluate to true and thus remain in our final sum for $\Pr(\varphi~\vert~do(a))$. By evaluating $\varphi$ first we may therefore greatly reduce the number of causal models that require evaluation under $\Pr$. This, alongside being able to factor out terms $\varphi(\textbf{POST})$ and $\Pr(\textbf{PRE})$ from $\delta_{a,a',\varphi}$, means that our actual computations will be far more efficient than this worst-case bound. 

Calculating the cost of an action $c(a)$ is a simple matter of summing over all possible outcomes in $\mathcal{O}$ and contexts in $\mathcal{X}$, evaluating the utility of each combination (complexity $O(U)$) and two probabilities (each having complexity $O(\vert P \vert )$): that of the context and that of the outcome given action $a$ and the context. As described above, once we have $\delta_{a,a',\varphi}$, $c(a)$, and $c(a')$, both $db_{N}(a, a', \varphi)$ and $db_{N}(a, \varphi)$ are computed as in HK, where $db_{N}(a, a', \varphi)$ requires the computation of $\delta_{a,a',\varphi}$ and the costs of two actions $c(a)$ and $c(a')$, and $db_{N}(a, \varphi)$ requires the same process $\vert\mathcal{R}(A)\vert$ times. Combining and factoring the complexity results for $\delta_{a,a',\varphi}$ and $c(a)$ accordingly gives us the time complexities for calculating blame using our embedding.

Finally, we observe that all of these complexities are exponential in the size of at least some subset of the variables. This is a result of the Boolean representation; our results are, in fact, \textit{more tightly bounded versions of those in HK}, which are each polynomial in the size of $\vert\mathcal{K}\vert = O(2^{\vert \mathcal{X} \vert + \vert \mathcal{D} \vert + \vert \mathcal{O} \vert})$. In practice, however, we only sum over worlds with non-zero probability of occurring. Using PSDDs allows us to exploit this fact in ways that other models cannot, as we can logically constrain the model to have zero probability on any impossible world. Thus, when calculating blameworthiness we can ignore a great many of the terms in each sum and speed up computation dramatically. To give some concrete examples, the model counts (variable assignments with non-zero probability) of the PSDDs in our three experiments were 52, 4800, and 180 out of $2^{12}$, $2^{21}$, and $2^{23}$ total variable assignments, respectively.

\section{Implementation}
\label{implementation}

\begin{figure}[H]
	\centering
	\begin{minipage}[c]{0.4\textwidth}
		\centering
		\includegraphics[width=\linewidth,trim={1cm 38.5cm 1cm 1cm},clip]{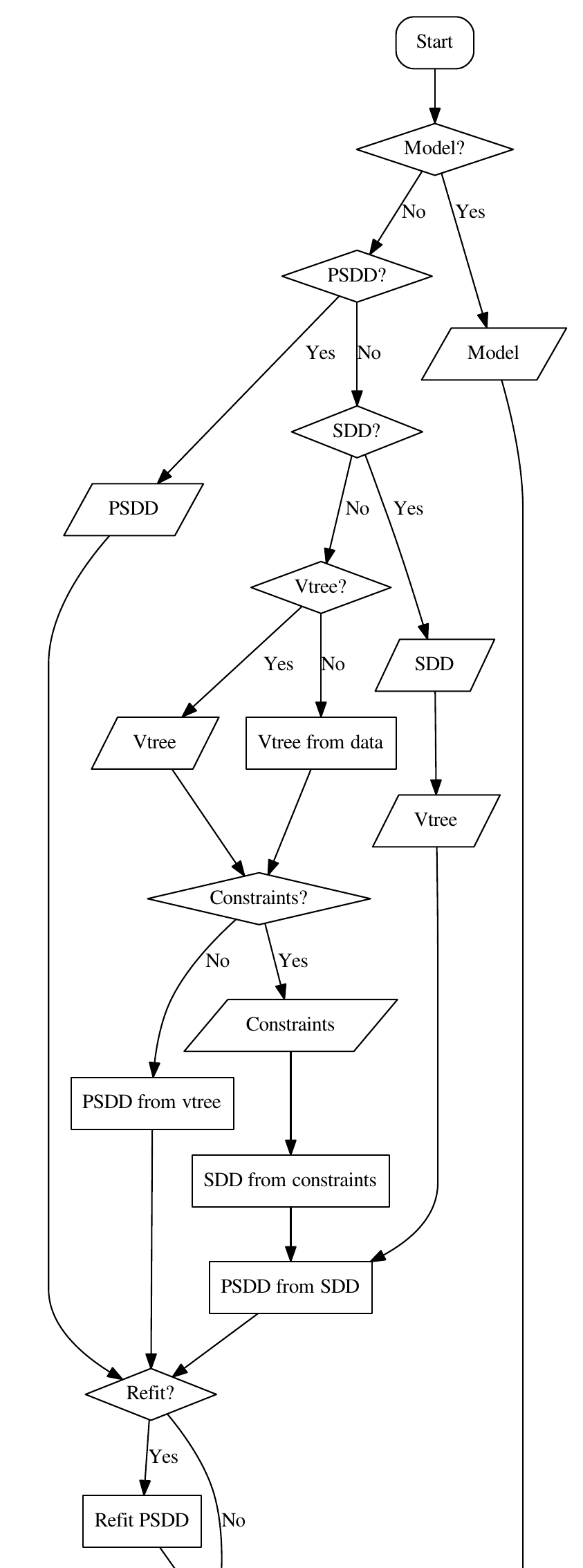}
		\label{fig:sub1}
	\end{minipage}
	\begin{minipage}[c]{0.4\textwidth}
		\centering
		\includegraphics[width=\linewidth, trim={1cm 1cm 1cm 39cm},clip]{flowchart.pdf}
		\label{fig:sub2}
	\end{minipage}
	\caption{The control flow of our system, split into two halves. Rounded rectangles are start and end points, diamonds represent decisions, parallelograms correspond to inputs from the user, and rectangles are processes undertaken by the program. \textit{q} refers to an optional alternative distribution over context and preceding decision/observation variables (allowing our model to re-used in a variety of scenarios) and \textit{N} refers to the measure of cost importance defined earlier.}
	\label{fig:pipeline}
\end{figure}

The importance of having \textit{implementable} models of moral reasoning has been stressed by Charisi et al. \cite{charisi2017towards}, amongst others. In this section we provide a brief summary of our implementation, before proceeding to experimentally evaluate it in Section \ref{experiments}. The underlying motivation behind our demonstration system (a high-level overview of which can be seen in Figure \ref{fig:pipeline}) was that a user should be able to go from any stage of creating a model to generating blameworthiness scores as conveniently and as straightforwardly as possible. Any inputs and outputs can be saved and thus each model and its results can be easily accessed and re-used if needed. Our implementation makes use of two existing resources: The SDD Package 2.0 \cite{SDD}, an open-source system for creating and managing SDDs, including compiling them from logical constraints; and LearnPSDD \cite{liang2017learning}, a recently developed set of algorithms that can be used to learn the parameters and structure of PSDDs from data, learn vtrees from data, and to convert SDDs into PSDDs. While this work is not the appropriate place to explain the precise details of this software (we refer the interested reader to \cite{darwiche2011sdd,SDD,liang2017learning}), we give a brief description of their workings in what follows.

The SDD Package 2.0 functions by initialising an SDD based on a vtree (which can be created at the same time or read from a file) and then gradually constructs the model from a propositional logic theory using a series of logical operations that are sequentially applied to larger and larger sub-SDDs over the set of variables. LearnPSDD can be used either with or without logical constraints. When used with constraints, the structure of the PSDD is found by compiling them into an SDD (as described above), and the parameters are given by their maximum likelihood estimates (possibly with smoothing). Without constraints, LearnPSDD first learns a vtree over a set of variables by splitting branches between variable subsets that minimise the average pairwise mutual information between the variables, then iteratively performs two operations, split and clone (described in more detail in \cite{liang2017learning}), on an initialised PSDD over the given vtree until a time or size limit is reached, or until the log likelihood of the model converges.

As well as making use of existing code, we also provide novel code for the remaining parts of the overall learning framework. These are as follows:

\begin{itemize}
	\item Building and managing models, and accepting various optional user inputs such as hand-specified utility functions or logical constraints specified in simple infix notation (e.g. $(A \land B) \leftrightarrow C$ can be entered using  \verb|=(&(A,B),C)|) and then converted to restrictions upon the learnt model. Being able to run the demonstration without using The SDD Package 2.0 or LearnPSDD directly greatly simplifies the interface to these two packages.
	\item Performing inference by evaluating the model or by calculating the most probable evidence (MPE), both possibly given partial evidence (as these functionalities are not provided in the original LearnPSDD package). Each of our inference algorithms are linear in the size of the model, and are based on pseudocode given in \cite{kisa2014probabilistic} and \cite{peharz2017latent} respectively.
	\item Learning utility functions from data, whose properties (such as being linear or being context-relative) are specified by the user in advance. This learning is done by forming a matrix equation representing our assumed proportionality relationship $\Pr(\textbf{D}~\vert~\textbf{X}) \propto f(\sum_{\textbf{O}}\mathrm{U}(\textbf{O})\Pr(\textbf{O}~\vert~\textbf{D},\textbf{X}))$ across all decisions and contexts, then solving to find utilities using non-negative linear regression with L2 regularisation (equivalent to solving a quadratic program). In particular, writing $A =  \Pr(\textbf{O}~\vert~\textbf{D},\textbf{X})$, $b =  \Pr(\textbf{D}~\vert~\textbf{X})$, and $x = \mathrm{U}(\textbf{O})$, we solve $\argmin_x(\Vert Ax-f^{-1}(b)\Vert^2_2 + \lambda\Vert x \Vert^2_2)$ where $\lambda$ is a regularisation constant and $\Vert \cdot \Vert~^2_2$ is the square of the Euclidean norm.
	\item Computing blameworthiness by efficiently calculating the key quantities defined by our embedding described in Section \ref{embedding}, using parameters for particular queries given by the user when required. Results are displayed in natural language and automatically saved for future reference.
\end{itemize}

The packaged version of our implementation (including full documentation), our data, and the results of our experiments detailed below are available online \cite{code}.

\section{Experiments and Results}
\label{experiments}

Using our implementation we learnt several models using a selection of datasets from varying domains in order to test our hypotheses. In particular we answer three questions in each case: 
\begin{itemize}[leftmargin=9mm]
    \item[(Q1)] \textit{Does our system learn the correct overall probability distribution?}
    \item[(Q2)] \textit{Does our system capture the correct utility function?}
    \item[(Q3)] \textit{Does our system produce reasonable blameworthiness scores?}
\end{itemize}
In this section we first summarise the results from our three experiments before providing a more in-depth analysis of our final experiment as an example. We direct the interested reader to Appendix \ref{otherexperiments} for results from the other two experiments. Appendix \ref{data} contains summaries of our datasets.

\subsection{Summary}

We performed experiments on data from three different domains. In \textit{Lung Cancer Staging} we used a synthetic dataset generated from the lung cancer staging influence diagram given in \cite{nease1997use}. The data was generated assuming that the overall medical decision strategy recommended in the original paper is followed with some high probability at each decision point. In this experiment, the utility of an outcome is captured by the expected length of life of the patient given that outcome, and the aim should be to make decisions regarding the diagnostic tests or treatments to apply that maximise this, meaning blame could reasonably be attributed to decisions that fail to do so. The \textit{Teamwork Management} experiment uses a recently collected dataset of human decision-making in teamwork management \cite{yu2017dataset}. This data was recorded from over 1000 participants as they played a game that simulates task allocation processes in a management environment, and includes self-reported emotional responses from each participant based on their performance. Here, different levels of blame are attributed to decision strategies that lead to lower-self reported happiness scores with respect to the various levels of the game and outcomes that measure performance such as the timeliness and quality of the work managed. Finally, in \textit{Trolley Problems} we devised our own experimental setup with human participants, using a small-scale survey (documents and data are included in the code package \cite{code}) to gather data about hypothetical moral decision-making scenarios. These scenarios took the form of non-deterministic and expanded variants on the famous trolley problem \cite{thomson1985trolley}, where blame can quite intuitively be attributed (as explained in more detail in Subsection \ref{trolley} below) to the participant's decisions about who should live or die depending on the context and outcomes.

For (Q1), we begin by noting that although we embed a causal framework in our choice of statistical relational model (PSDDs), that (as shown in Section \ref{probz}) the causal queries we need to answer within this framework can be computed using standard probabilistic methods. Thus, the question of how well we are able to answer such queries reduces to the question of how well we are able to compute the relevant probabilities, and thus to how well our system learns the correct overall probability distribution. This essentially requires an evaluation of density estimation, which we measure via the overall log likelihood of the models learnt by our system on training, validation, and test datasets (see Table \ref{tab:modelcomp}). A full comparison across a range of similar models and learning techniques is beyond the scope of our work here, although to provide some evidence of the competitiveness of PSDDs we include the log likelihood scores of a sum-product network (SPN), another tractable probabilistic model, created using Tachyon \cite{tachyon} as a benchmark. We also compare the sizes (measured by the number of nodes) and the log likelihoods of PSDDs learnt with and without logical constraints in order to demonstrate the effectiveness of the former approach. We reiterate here that we include these comparisons not to thoroughly benchmark our models against a suite of baselines, but merely to indicate that their performance is in line with competitors. A brief further discussion of said competitors and related models in probabilistic logic learning is included in Section \ref{related}. In Section \ref{trolley} we also provide, as an illustrative example, a more intuitive visual representation of the learnt marginal distribution over decision variables for one particular moral decision-making scenario.

\begin{table}[H]
\begin{center}
\begin{tabular}{c l l l l l}
\toprule
\multicolumn{2}{l }{\textbf{Model}} & \textbf{Training} & \textbf{Validation} & \textbf{Test} & \textbf{Size}\\
        \midrule
		& PSDD* & -2.047 & -2.046 & -2.063 & 134\\
		\textbf{1}& PSDD & -2.550 & -2.549 & -2.564 & 436\\
		& SPN & -3.139 & -3.143 & -3.158 & 1430\\
		\midrule
		& PSDD* & -5.541 & -5.507 & -5.457 & 370\\
		\textbf{2}& PSDD & -5.637 & -5.619 & -5.556 & 931\\
		& SPN & -7.734 & -7.708 & -7.658 & 3550\\
		\midrule
		& PSDD* & -4.440 & -4.510 & -4.785 & 368\\
		\textbf{3}& PSDD & -6.189 & -6.014 & -6.529 & 511\\
		& SPN & -15.513 & -16.043 & -15.765 & 3207\\
		\bottomrule
\end{tabular}
\end{center}
\caption{Log-likelihoods and sizes of the constrained PSDDs (the models we use in our system, indicated by the * symbol), unconstrained PSDDs, and the SPNs learnt in our three experiments. Higher log-likelihoods are better, as are lower sizes (measured by the number of model parameters).}
\label{tab:modelcomp}
\end{table}

Answering (Q2) is more difficult, as self-reported measures of utility (or other proxy metrics, such as life expectancy in \textit{Lung Cancer Staging}, for example) may form an unreliable baseline. More generally, one might argue that to attempt to measure utility quantitatively is problematic in and of itself. Though discussion of this question is beyond the scope of our work here, we note that in recent years, with experiments such as the `Moral Machine' \cite{MM}, we have seen efforts to capture the moral judgements of humans in a principled and quantitative fashion. It is also the case that in many applications (such as the use of QALYs in healthcare, or the field of preference elicitation), things of moral value are evaluated using a quantitative framework in a way that is widely accepted by professionals in that area as well as by moral philosophers. In our experiments, our models are able to learn utility functions that match preferences up to ordinality in most cases, but the cardinal representations of utilities depends greatly on the function $f$ in the proportionality relationship between expected decision probabilities and expected utilities. The exact choice of $f$ is highly domain-dependent and an area for further experimentation in future.

In attempting to answer (Q3) we divide our question into two parts: does the system attribute no blame in the correct cases, and does the system attribute more blame in the cases we would expect it to (and less in others)? Needless to say, similar concerns such as those raised above about the measurement of utility apply to the notion of blame, and it is very difficult (perhaps even impossible, at least without an extensive survey of human opinions) to produce an appropriate metric for how correct our attributions of blame are. However, we suggest that these two criteria are the most fundamental and capture the core of what we want to evaluate in these initial experiments. We successfully queried our models in a variety of settings corresponding to the two questions above and present representative examples below.

\subsection{Trolley Problems}
\label{trolley}

\begin{figure}[H]
	\centering
	\includegraphics[width=0.8\textwidth,trim={0.5cm 0.5cm 0.5cm 0.5cm},clip]{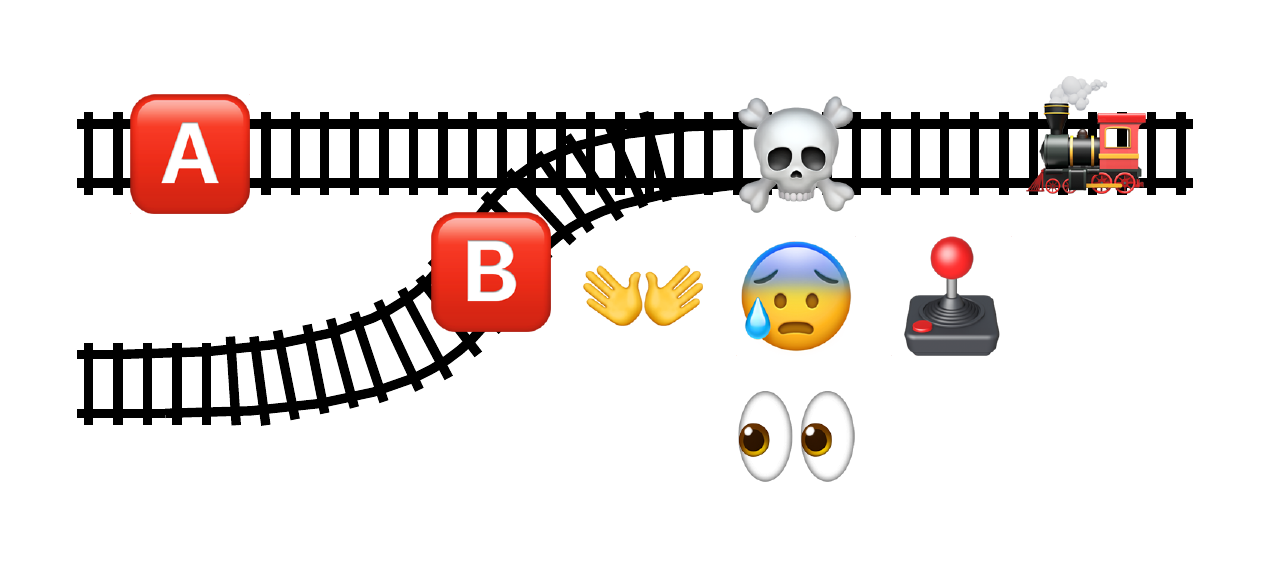}
	\caption{A cartoon given to participants showing the layout of the experimental scenario and the four possible options. Clockwise from top (surrounding the face symbol) these are: sacrificing oneself, flipping the switch, inaction, and pushing the character at $B$ onto the main track. Locations $A$ and $B$ are instantiated by particular characters depending on the context.}
	\label{fig:experiment}
\end{figure}

In this experiment we extend the well-known trolley problem, as is not uncommon in the literature \cite{MM}, by introducing a series of different characters that might be on either track: one person, five people, 100 people, one's pet, one's best friend, and one's family. We also add two further decision options: pushing whoever is on the side track into the way of the train in order to save whoever is on the main track, and sacrificing oneself by jumping in front of the train, saving both characters in the process. Our survey then took the form of asking each participant which of the four actions they would perform (the fourth being inaction) given each possible permutation of the six characters on the main and side tracks (we assume that a character could not appear on both tracks in the same scenario). The general setup can be seen in Figure \ref{fig:experiment}, with locations $A$ and $B$ denoting the locations of people on the main track and side track respectively. 

Last of all, we added a probabilistic element to our scenarios whereby the switch only works with probability 0.6, and pushing the character at location $B$ onto the main track in order to stop the train succeeds with probability 0.8. This was used to account for the fact that people are generally more averse to actively pushing someone than to flipping a switch \cite{singer2005ethics}, and people are certainly more averse to sacrificing themselves than doing either of the former. However, depending on how much one values the character on the main track's life, one might be prepared to perform a less desirable action in order to increase their chance of survival.

\begin{figure}[H]
	\centering
	\includegraphics[width=0.8\textwidth,trim={2cm 12.5cm 2cm 7cm},clip]{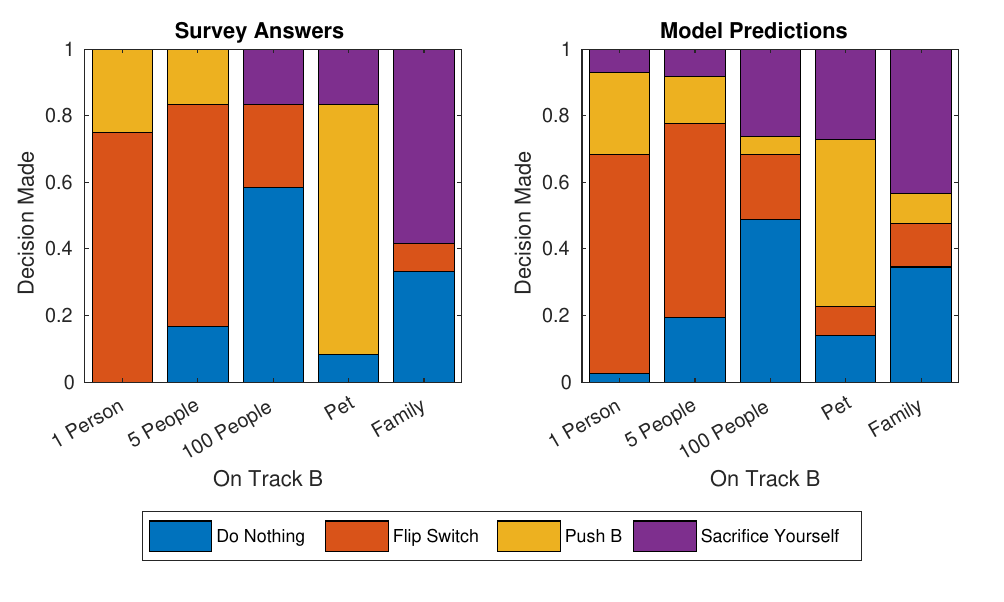}
	\caption{A comparison of the decisions made by participants and the predictions of our model in each of the five scenarios in which the best friend character is on the main track ($A$).}
	\label{fig:trolleycompare}
\end{figure}

In answering (Q1), as well as the primary log-likelihood metric recorded in Table \ref{tab:modelcomp}, for illustrative purposes we also investigate how well our model serves as a representation of the aggregated decision preferences of participants by calculating how likely the system would be to make particular decisions in each of the 30 contexts and comparing this with the average across participants in the survey. For reasons of space we focus here on a representative subset of these comparisons: namely, the five possible scenarios in which the best friend character is on the main track (see Figure \ref{fig:trolleycompare}). In general, the model's predictions are similar to the answers given in the survey, although the effect of smoothing our distribution during learning is noticeable, especially due to the fact that the model was learnt with relatively few data points. Despite this handicap, the most likely decision in any of the 30 contexts according to the model is in fact the majority decision in the survey, with the ranking of other decisions in each context also highly accurate. 

Unlike our other two experiments, the survey data does not explicitly contain any utility information, meaning our system was forced to learn a utility function by using the probability distribution encoded by the PSDD. Within the decision-making scenarios we presented, it is plausible that the decisions made by participants were guided by weights that they assigned to the lives of each of the six characters and to their own life. Given that each of these is captured by a particular outcome variable we chose to construct a utility function that was linear in said variables. We also chose to make the utility function insensitive to context, as we would not expect how much one values the life of a particular character to depend on which track that character was on, or whether they were on a track at all.

\begin{figure}[H]
	\centering
	\includegraphics[width=0.8\textwidth,trim={2.5cm 9cm 2.5cm 10cm},clip]{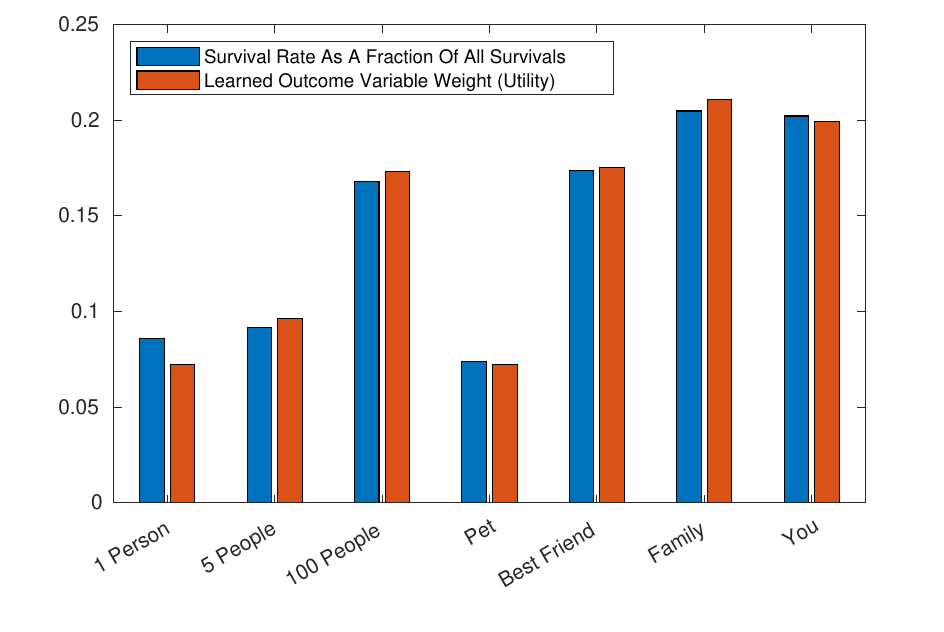}
	\caption{A comparison between the average survival rates of the seven characters (including the participants in the survey), normalised to sum to one, and the corresponding utility function weights learnt by our system.}
	\label{fig:trolleycompareutil}
\end{figure}

For (Q2), with no existing utility data to compare our learnt function, we interpreted the survival rates of each character as the approximate weight assigned to their lives by the participants. While the survival rate is a non-deterministic function of the decisions made in each context, we assume that over the experiment these rates average out enough for us to make a meaningful comparison with the weights learnt by our model. A visual representation of this comparison can be seen in Figure \ref{fig:trolleycompareutil}. It is immediately obvious that our system has captured the correct utility function to a high degree of accuracy. With that said, our assumption about using survival rates as a proxy for real utility weights does lend itself to favourable comparison with a utility function learnt from a probability distribution over contexts, decisions, and outcomes (which therefore includes survival rates). Given the setup of the experiment, however, this assumption seems justified and, furthermore, to be in line with how most of the participants answered the survey.

Because of the symmetric nature of the set of contexts in our experiment, the probability of a particular character surviving as a result of a particular fixed action across all contexts is just the same as the probability of that character not surviving. Hence in answering (Q3) we use our system's feature of being able to accept particular distributions $\Pr'$ over the contexts in which we wish to attribute blame, allowing us to focus only on particular scenarios. Regarding the first part of (Q3), clearly in any of the possible contexts one should not be blamed at all for the death of the character on the main track for flipping the switch ($F$) as opposed to inaction ($I$), because in the latter case they will die with certainty, but not in the former.\footnote{Note that this is not to say one would not be blameworthy when compared to all other actions as one could, for example, have sacrificed oneself instead, saving all other lives with certainty.} Choosing a scenario arbitrarily to illustrate this point, with one person on the side track and five people on the main track, we have $db_N(F, I, \neg L_{5}) = 0$ and $db_N(F, \neg L_{5}) = 0.307$ (with our measure of cost importance $N = 0.762$, 1.1 times the negative minimum cost of any action).

For the second part of (Q3), consider the scenario in which there is a large crowd of a hundred or so people on the main track, but one is unable to tell from a distance if the five or so people on the side track are strangers or one's family. The more likely it is that the family is on the side track, the more responsible one is for their deaths ($\neg L_{Fa}$) if one, say, flips the switch ($F$) to divert the train. Conversely, we also expect there to be \textit{less} blame for the deaths of the 100 people ($\neg L_{100}$) say, if one did nothing ($I$), the more likely it is that the family is on the side track (because the cost, for the participant at least, of diverting the train is higher). We compare cases where there is a 0.3 or 0.6 probability that the family is on the side track and for all calculations use the cost importance measure $N = 1$. Therefore, not only would we expect the blame for the death of the family to be higher when pulling the switch in the latter case, we would expect the value to be approximately twice as high as in the former case. Accordingly, we compute values $db_N(F, \neg L_{Fa}) = 0.264$ and $db_N(F, \neg L_{Fa}) = 0.554$ respectively. Similarly, when considering blame for the deaths of the 100 people due inaction, we find that $db_N(I, \neg L_{100}) = 0.153$ in the former case and that $db_N(I, \neg L_{100}) = 0.110$ in the latter case (when the cost of performing another action is higher).

\section{Discussion}
\label{discussion}

We begin this section by briefly revisiting two of the technical points in Section \ref{varz} and Section \ref{probz}: a) structural equations can be partially encoded in PSDDs using propositional formulae; and b) the specific sorts of causal queries we make in our framework can be reduced to a number of probabilistic queries. In particular, we wish to highlight the fact that these claims are \emph{independent} and used to support independent arguments. The first claim supports our argument that PSDDs are a natural choice of model due to (amongst other features) their relation to causal graphical models and the structural equations that they represent (when compared to, say, alternatives such as SPNs \cite{Papantonis2019}). The second claim supports our argument that PSDDs (or probabilistic models more generally) are sufficient to answer the particular set of causal queries within the particular class of sequential decision-making scenarios we consider, and are thus an appropriate choice of model in which to embed the formal framework of HK. Further, the truth of the second claim is what justifies our focus on computing probabilistic quantities instead of the process of causal discovery which, as noted earlier, is highly non-trivial. Admittedly, such learning regimes would be interesting and useful in our context, and we plan to look into this in future work. 

As well the technical assumptions discussed in Section \ref{embedding}, our work also rests on several key philosophical assumptions worthy of discussion. These are in turn linked to our motivations and suggestions for potential applications of the type of system we exhibit. We discuss each of these aspects with respect to the features and abilities of our system below.

Most importantly, we wish to draw attention to what we consider an interesting parallel between the use of statistical relational models that can encode both logical constraints or structures as well as learnt distributions (which can in turn be used to deduce preferences), and normative ethical theories that make use of some notion of both deontological rules (e.g. it is forbidden to kill another human being) and the principle of utility maximisation. While these two philosophical approaches are often contrasted with each other, it is plausible and not infrequently suggested that human beings make use of both in their everyday moral reasoning \cite{conway2013deontological}. For example, this helps to explain why many people consider it morally permissible to flip a switch to kill one person and save five, but not to push someone to their death in order to save five others (as it would violate a deontological rule forbidding killing that is not violated in the first, more `indirect' case) \cite{singer2005ethics}. This parallel suggests that such models (including PSDDs) may have an intrinsic advantage when it comes to capturing the complexities of moral reasoning. It is perhaps also possible that biased data used for learning could be more easily identified (through the use of complex logical queries) or perhaps restricted (through the use of logical constraints) by these models, though this is of course a highly non-trivial problem.

With respect to our specific embedding and implementation, we can easily constrain our distribution and thus the utility function that results (for example, in the trolley problems experiment we could have encoded logical constraints such that any human life should be prioritised over the life of a pet). Bounding our models before learning in this way corresponds to a hybrid between the top-down and bottom-up approaches defined by Allen et al. which we believe seems intuitively more promising and flexible than using either technique exclusively \cite{allen2005artificial}. The possibility of such a hybrid system incorporating both statistical and symbolic methods has also been discussed elsewhere \cite{charisi2017towards}, though as far as we are aware our system represents the first implemented example of this paradigm. A less immediate but more general feature is our ability to tractably query an unconstrained model in order to check with what probability certain rules are followed, based on contexts and (possibly) previous decisions.

Though the primary purpose of our models is in \textit{representing} moral decision-making scenarios for tractable \textit{reasoning} about decisions, outcomes, and blame, they can also be used to \textit{make} such decisions tractably, using our implemented MPE algorithm. However, we do not wish to suggest blindly advocating the automation of moral judgements. In our view, it is crucial that AI systems act in line with human values and preferences. Our suggestion in this work is merely that imbuing such systems with the ability to reason about moral value, blame, and intentionality is one possible step towards this goal.\footnote{It is a separate but potentially interesting question to ask whether a group of purely artificial agents might benefit (say, in their level of coordination) from such abilities.} Our motivation derives from HK (and others) in our desire to provide \textit{a shared computational framework for representing and reasoning about moral judgements} that may help in our quest to build systems that act ethically; the difference being that we contribute a concrete, end-to-end implementation and investigation of such a framework as opposed to an underlying logical theory. 

As autonomous systems become more widely and deeply embedded within society, and as the quantity and significance of their interactions with (or on behalf of) humans grows, so too, we believe, will the need for a computationally realised framework of the kind we present here \cite{conitzer2017moral,moor2006nature,charisi2017towards}, whether or not it is used to make or merely reason about moral decisions. We wish to remark, however, that if this framework were employed in the wrong way, such as in the unchecked automation of moral decision-making tasks, then it could undoubtedly lead to unethical consequences (see, for example, \cite{asaro2012banning} in opposition to autonomous weapons systems). With that said, it would also be na\"{\i}ve to think that the decisions made by current and future autonomous systems are without moral consequence, and so the important discussion surrounding these issues is one that we believe will undoubtedly continue and hope to encourage through our work here.

Though there are many other related ethical considerations that warrant discussion, a detailed investigation of such issues is outside the scope of our current work, and so we conclude this section with suggestions for possible applications of our work (or extensions thereof). Beginning with our three experiments: the first represents a case in which, after learning from previous expert behaviour or having certain parameters specified in advance, a system like ours could, for example, be used to quantify culpability in the event of a patient's death due to medical error; in the second experiment we could use a similar model for the process of After Action Review within a team training setting (as proposed in \cite{gratch2003automating}); and the models extracted during our third experiment could be used for comparison against learnt models from specific individuals or other populations, and potentially also what Etzioni and Etzioni describe as `ethics bots': personalised models learnt from data that encode moral preferences and may be transferred between domains \cite{etzioni2017incorporating}.

In addition, autonomous systems that can reason accurately and tractably about blame and moral responsibility could see use in ensuring politeness in natural language generation \cite{briggs2014modeling}, creating shared understanding in collaborative tasks between multiple human and/or artificial agents (though see \cite{groom2010critic,kaniarasu2014effects} for possible negative side effects of autonomous systems blaming humans), overcoming human cognitive biases in legal, military, or other high-stakes scenarios \cite{arkin2009ethical}, and many others.

\section{Related Work}
\label{related}

Our work here is differentiated from related work in two main ways: jointly addressing \emph{the  automated  learning of models of moral scenarios}, and \emph{tractable reasoning}. We discuss other efforts below.

As mentioned before, we do not motivate new definitions for moral responsibility here but draw on HK which, in turn, is based upon prior work done by Halpern with Chockler \cite{chockler2004responsibility} and with Pearl \cite{halpern2005causes}. Their framework is also related to the intentions model of Kleiman-Weiner et al. which considers predictions about the moral permissibility of actions via influence diagrams \cite{kleiman2015inference}, though unlike our efforts here all of these works are primarily theoretical and there is no emphasis on learning or tractability. In fact, the use of tractable architectures for decision-making itself is recent (see, for example, \cite{bhattacharjya2012evaluating,melibari2016sum}). Choi et al. learn PSDDs over preference rankings (as opposed to decision-making scenarios more generally) \cite{choi2015tractable}, though unlike ours their approach does not take account of different preferences in different contexts and does not capture the causal elements we adopt from HK.

Just as the focus of this work is not to provide a new definition of moral responsibility, it is equally not to introduce a new tractable probabilistic architecture. Instead, we adopt PSDDs which offer a useful combination of learning in the presence of logical (and therefore possibly moral) constraints and then tractably computing the many quantities needed for the blameworthiness framework of HK. With that said, this does not mean that other models could not have been used. In principle, any tractable fragment from probabilistic logic learning is perhaps applicable to our work here: we initially considered a decision-theoretic instance of SPNs \cite{melibari2016sum}, but chose not to pursue this further due to the focus on \textit{making} decisions as opposed to \textit{reasoning about} decisions, and the lack of an available code base. We could also perhaps have leveraged a high-level language like DTProbLog \cite{dtproblog}, but we note that there is the exponential cost of compiling such a language to a circuit, and thus we wished to work with the circuit directly. 

Markov Logic Networks (MLNs) could have been considered, and when their semantics are viewed from the perspective of weighted model counting, they are equivalent to the task solved by probabilistic circuits \cite{mlns}. Analogously, a tractable fragment of MLNs could have been considered, as could Probabilistic Soft Logic which supports convex optimisation during inference but a fuzzy/t-norm type semantics \cite{psl}. The main challenge in extending our work for any of these other proposal languages would be identifying an embedding from HK to the target language, but once that is resolved, we would expect to see similar results (insofar as the models support the class of queries required in order to tractably compute blameworthiness). Thus, we do not claim that PSDDs are the only route to the contributions in this work, though their tractable nature, as well as the partial encoding of the structural equations through the use of logical constraints, allow a clean practical perspective on HK, and coupled with parameter estimation and utility learning, we obtain the corresponding implemented framework.

An important part of learning a model of moral decision-making is in learning a utility function. This is often referred to  as \textit{inverse reinforcement learning} \cite{ng2000algorithms} or \textit{Bayesian inverse planning} \cite{baker2009action} and is closely related to the field of preference elicitation. Our current implementation considers a simple approach for learning utilities (similar to that of Nielsen and Jensen \cite{nielsen2004learning}), but more involved paradigms such as those above could indeed have been used. Existing work in these areas, however, typically has the extraction of utilities as a final goal, whereas in our work such utilities are merely inputs for moral reasoning processes. Learning preferences from sub-optimal behaviour is an important challenge here that we hope to take into account in future work. We refer the interested reader to the work of Evans et al. for details of one attempt to tackle this problem \cite{evans2016learning}. Recent work by Jentzsch et al. indicates that language corpora may form suitable resources from which data about ethical norms and moral decision-making may be extracted, which may help in our ability to learn larger and more complex models in future \cite{Jentzsch}.

Our contributions here are related to the body of work surrounding  MIT's Moral Machine project \cite{MM}. For example, Kim et al. \cite{kim2018computational} build on an earlier theoretical proposal \cite{kleiman2017learning} by developing a computational model of moral decision-making whose predictions they test against Moral Machine data. Their focus is slightly different to ours, as they attempt to learn abstract moral principles via hierarchical Bayesian inference. Although our framework can be used to these ends, it is also flexible with respect to different contexts, and allows constraints on learnt models. Noothigattu et al. develop a method of aggregating the preferences of all participants (this, while not using techniques that are quite as sophisticated, is a secondary feature of our system which is strictly more general) in order to make a given decision \cite{noothigattu2017voting}. However, due to the large numbers of such preference orderings, tractability issues arise and so sampling must be used. In contrast, inference in our models is both exact and tractable.

There have also been many purely symbolic approaches to creating models of moral reasoning within autonomous systems. As in our own work, the HERA project \cite{lindner2017hera} is also based on Halpern and Pearl's structural equations framework for causal and counterfactual reasoning \cite{halpern2005causes}. The system is similarly broad in that it allows for the implementation of several kinds of (rule-based) moral theory to be captured, however their models and utility functions are hand-crafted (as opposed to learnt from data) and so lack the flexibility, tractability, and scalability of our approach. Mao and Gratch also make use of causal models to produce categorical judgements of moral responsibility based on psychological attribution theory \cite{mao2012modeling}. While their focus is on the multi-agent setting (and again, does not include any learning), reasoning in such domains is also supported by the underlying theory of HK and would thus form a natural extension of our work in the future.

GENETH uses inductive logic programming to create generalised moral principles from the judgements of ethicists about particular ethical dilemmas, the system's performance being evaluated using an `ethical Turing test' \cite{anderson2014geneth}. This work, however, can be seen as less general than our approach in that they assume preferences to be ordinal (as opposed to cardinal) and actions to be deterministic (as opposed to probabilistic). Their need for feature engineering to extract ethically relevant facts from each situation is also bypassed by our system, though it is plausible that adding such variables to our models could improve results. Further symbolic approaches (as opposed to our own which forms a hybrid between symbolic and statistical methods) include the ETHAN language, the properties of agents defined by which are also amendable to formal verification \cite{dennis2016formal}, and work on simulating then evaluating the moral consequences of a robot's actions as part of its `ethical control layer' \cite{vanderelst2018architecture}.

Finally, there are a number of works that aim to provide overviews of or motivations for broad classes of algorithms that seek to address similar problems to those in our own work, though given their nature these works focus more on breadth than the deeper analysis of a single framework which we provide here. A discussion of strategies for creating moral decision-making frameworks for autonomous systems is discussed in Conitzer et al. \cite{conitzer2017moral}, and similar considerations regarding hybrid collective decision-making systems are made by Greene et al. \cite{greene2016embedding}. One alternative proposal, not discussed in either of these works, is made by Abel et al. and suggests the use of reinforcement learning as a framework for ethical decision-making \cite{abel2016reinforcement}. Recent work by Shaw et al. \cite{Shaw} has sought to address the tension between learnt models of moral decision-making and provable guarantees, in a work not dissimilar to our own. A comprehensive survey of issues surrounding the intersection of ethics and autonomous systems is provided by Charisi et al. \cite{charisi2017towards}. We refer the reader to these works for more discussions.

\section{Conclusion}
\label{conclusion}

In this work we present the first implemented hybrid (between data-driven and rule-based methods) computational framework for moral reasoning, which utilises the specification of decision-making scenarios in HK, and at the same time exploits many of the desirable properties of PSDDs (such as tractability, semantically meaningful parameters, and the ability to be both learnt from data and include logical constraints). The implemented system is flexible in its usage, allowing various inputs and specifications. In general, the models in our experiments are accurate representations of the distributions over the moral scenarios that they are learnt from. Our learnt utility functions, while simple in nature, are still able to capture subtle details and in some scenarios are able to match human preferences with high accuracy using very little data. With these two elements we are able to generate blameworthiness scores that are, prima facie, in line with human intuitions. 

We hope that our work here goes some way towards bridging the gap between the existing philosophical work on moral responsibility and the existing technical work on reasoning about decision-making in autonomous systems. In future we would like to expand the application of our implementation to more complex domains in order to fully exploit and evaluate its tractability. We are also interested in investigating how intentionality can be modelled within our embedding (a natural extension to the work presented here, given the close connection between this concept and blameworthiness), and the possibility of formally verifying certain properties of our models.

\section*{Acknowledgements}

Vaishak Belle was supported by a Royal Society University Research Fellowship. The authors wish to thank Yitao Liang, Giannis Papantonis, Craig Innes, Michael Varley, and Sophia Jones for useful discussions while conducting this work, as well as several anonymous reviewers whose comments and suggestions have helped to significantly improve this paper.

\bibliography{bibliography}

\renewcommand\thesection{\Alph{section}}
\setcounter{section}{0}

\section{Further Experiments}
\label{otherexperiments}

As well as the \textit{Trolley Problems} experiment in Section \ref{trolley}, we also applied our framework to data from moral decision-making scenarios in two other illustrative domains: \textit{Lung Cancer Staging} and \textit{Teamwork Management}. In the following subsections of this appendix we provide the details and results of these experiments.

\subsection{Lung Cancer Staging}

We use a synthetic dataset generated with the lung cancer staging influence diagram given in \cite{nease1997use}. The data was generated assuming that the overall decision strategy recommended in the original paper is followed with some high probability at each decision point. In this strategy, a thoractomy is the usual treatment unless the patient has mediastinal metastases, in which case a thoractomy will not result in greater life expectancy than the lower risk option of radiation therapy, which is then the preferred treatment. The first decision made is whether a CT scan should be performed to test for mediastinal metastases, the second is whether to perform a mediastinoscopy. If the CT scan results are positive for mediastinal metastases then a mediastinoscopy is usually recommended in order to provide a second check, but if the CT scan result is negative then a mediastinoscopy is not seen as worth the extra risk involved in the operation. Possible outcomes are determined by variables that indicate whether the patient survives the diagnosis procedure and survives the treatment, and utility is measured by life expectancy.

\begin{figure}[H]
	\centering
	\includegraphics[width=0.8\textwidth,trim={2.5cm 11cm 2.5cm 11cm},clip]{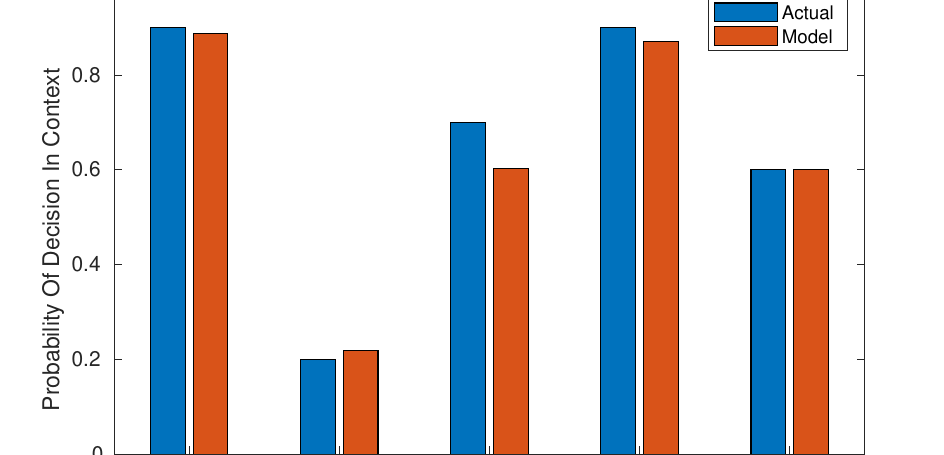}
	\caption{A comparison between the five probability values specified in our data generation process and the corresponding values learnt by our system from this data.}
	\label{fig:cancerstratprobs}
\end{figure}

For (Q1) we again measure the overall log likelihood of the models learnt by our system on training, validation, and test datasets. In particular, our model is able to recover the artificial decision-making strategy well (see Figure \ref{fig:cancerstratprobs}); at most points of the staging procedure the model learns a very similar distribution over decisions, and in all cases the correct decision is made the majority of times.

Answering (Q2) here is more difficult as the given utilities are not necessarily such that our decisions are linearly proportional to the expected utility of that decision. However, our strategy was chosen so as to maximise expected utility in the majority of cases. Thus, when comparing the given life expectancies with the learnt utility function, we still expect the same ordinality of utility values, even if not the same cardinality. In particular, our function assigns maximal utility (1.000) to the successful performing of a thoractomy when the patient does not have mediastinal metastases (the optimal scenario), and any scenario in which the patient dies has markedly lower utility (mean value 0.134).

Regarding the first part of (Q3), one case in which we have blameworthiness scores of zero is when performing the action being judged is \textit{less} likely to result in the outcome we are concerned with than the action(s) we are comparing it to. The chance of the patient dying in the diagnostic process ($\neg S_{DP}$) is increased if a mediastinoscopy ($M$) is performed, hence the blameworthiness for such a death due to \textit{not} performing a mediastinoscopy should be zero. As expected, our model assigns $db_N(\neg M, M, \neg S_{DP}) = 0$. To answer the second part of (Q3), we show that the system produces higher blameworthiness scores when a negative outcome is more likely to occur (assuming the actions being compared have relatively similar costs). For example, in the case where the patient does not have mediastinal metastases then the best treatment is a thoractomy, but a thoractomy will not be performed if the result of the last diagnostic test performed is positive. The specificity of a mediastinoscopy is  higher than that of a CT scan, hence a CT scan is more likely to produce a false positive and thus (assuming no mediastinoscopy is performed as a second check) lead to the wrong treatment.\footnote{Note that even though a mediastinoscopy has a higher cost (as the patient is more likely to die if it is performed), it should not be enough to outweigh the test's accuracy in this circumstance.} In the case where only one diagnostic procedure is performed we therefore have a higher degree of blame attributed to the decision to conduct a CT scan (0.013) as opposed to a mediastinoscopy (0.000), where we use $N = 1$.

\subsection{Teamwork Management}

Our second experiment uses a recently collected dataset of human decision-making in teamwork management \cite{yu2017dataset}. This data was recorded from over 1000 participants as they played a game that simulates task allocation processes in a management environment. In each level of the game the player has different tasks to allocate to a group of virtual workers that have different attributes and capabilities. The tasks vary in difficulty, value, and time requirements, and the player gains feedback from the virtual workers as tasks are completed. At the end of the level the player receives a score based on the quality and timeliness of their work. Finally, the player is asked to record their emotional response to the result of the game in terms of scores corresponding to six basic emotions. We simplify matters slightly by considering only the self-declared management strategy of the player as our decisions. Within the game this is recorded by five check-boxes at the end of the level that are not mutually exclusive, giving 32 possible overall strategies. These strategy choices concern methods of task allocation such as load-balancing (keeping each worker's workload roughly even) and skill-based (assigning tasks by how likely the worker is to complete the task well and on time), amongst others. We also measure utility purely by the self-reported happiness of the player, rather than any other emotions.

\begin{figure}[H]
	\centering
	\includegraphics[width=0.8\textwidth,trim={2.5cm 9cm 2.5cm 10cm},clip]{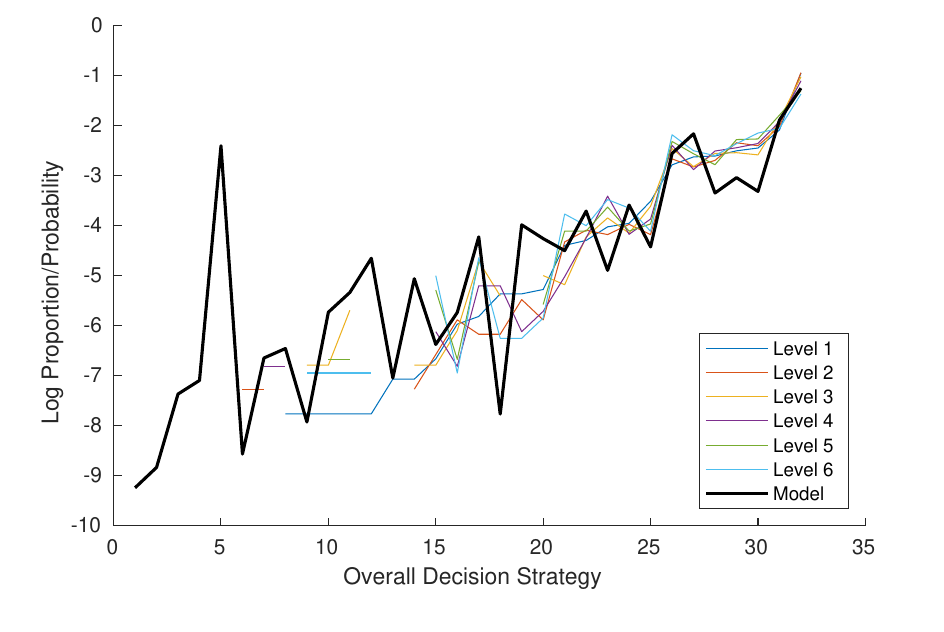}
	\caption{The log probability assigned to each possible decision strategy across all contexts by our model, compared to the log proportion of times each strategy was used in the six levels of the game by participants. Strategies are sorted in ascending order by their proportion of use in level 1 and gaps in each plot represent strategies never used in that game level.}
	\label{fig:teamworkdecs}
\end{figure}

As part of our answer to (Q1) we investigate how often the model would employ each of the 32 possible strategies (where a strategy is represented by an assignment of values to the binary indicator decision variables) compared to the average participant (across all contexts), which can be seen in Figure \ref{fig:teamworkdecs}. In general the learnt probabilities are similar to the actual proportions in the data, though noisier. The discrepancies are more noticeable (though understandably so) for decisions that were made very rarely, perhaps only once or twice in the entire dataset. These differences are also partly due to smoothing (i.e. all strategies have a non-zero probability of being played).

\begin{figure}[h]
	\centering
	\includegraphics[width=0.8\textwidth,trim={2.5cm 9cm 2.5cm 10cm},clip]{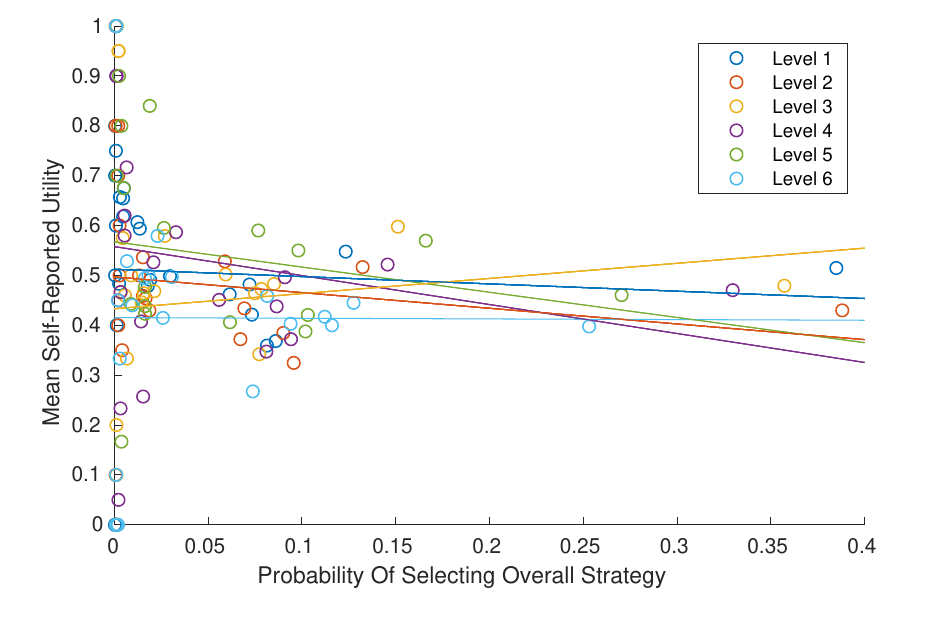}
	\caption{Each point is a decision strategy in a level of the game; we compare the proportion of times it is used against the average self-reported utility that results from it. Each line is a least-squares best fit to the points in that level.}
	\label{fig:teamworkscatter}
\end{figure}

For (Q2) we use the self-reported happiness scores to investigate our assumption that the number of times a decision is made is (linearly) proportional to the expected utility based on that decision. In order to do this we split the data up based on the context (game level) and produce a scatter plot (Figure \ref{fig:teamworkscatter}) of the proportion of times a set of decisions is made against the average utility (happiness score) of that decision. Overall there is no obvious positive linear correlation as our original assumption would imply, although this could be because of any one or combination of the following reasons: players do not play enough rounds of the game to find out which strategies reliably lead to higher scores and thus (presumably) higher utilities; players do not accurately self-report their strategies; or players' strategies have relatively little impact on their overall utility based on the result of the game. We recall here that our assumption essentially comes down to supposing that people more often make decisions that result in greater utilities. The eminent plausibility of this statement, along with the relatively high likelihood of at least one of the factors in the list above means we do not have enough evidence here to refute the statement, although certainly further empirical work is required in order to demonstrate its truth.

\begin{figure}[H]
	\centering
	\includegraphics[width=0.8\textwidth,trim={2.2cm 14cm 2cm 6cm},clip]{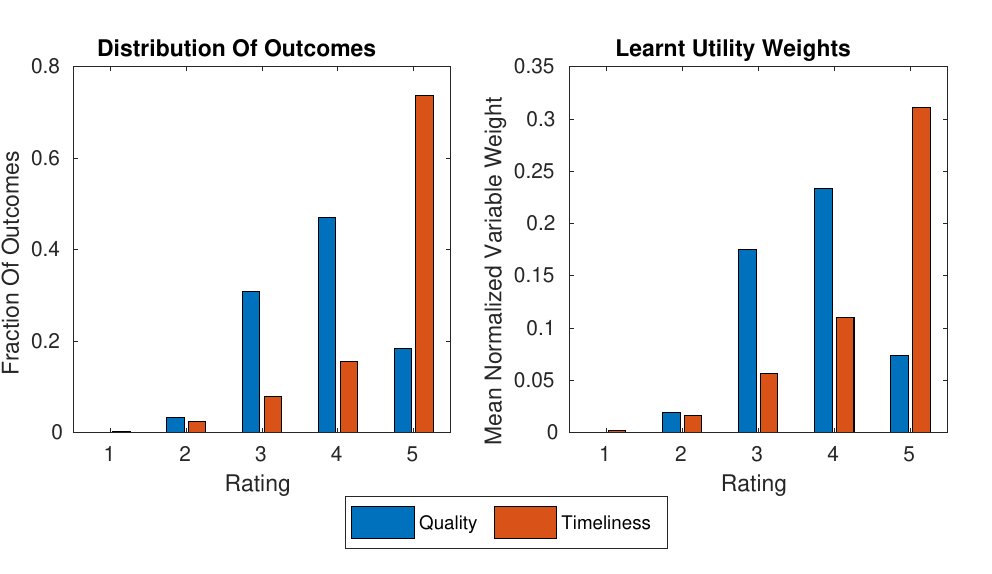}
	\caption{A comparison of the learnt utility weights for each of the outcome variables (to the right) and the proportion of times each outcome occurs in the data (to the left).}
	\label{fig:teamworkweights}
\end{figure}

Investigating this discrepancy further, we learnt a utility function (linear and context-relative) from the data and inspected the average weights given to the outcome variables (see right plot in Figure \ref{fig:teamworkweights}). A correct function should place higher weights on the outcome variables corresponding to higher ratings, which is true for timeliness, but not quite true for quality as the top rating is weighted only third highest. We found that the learnt utility weights are in fact almost identical to the distribution of the outcomes in the data (see left plot in Figure \ref{fig:teamworkweights}). Because our utility weights were learnt on the assumption that players more often use strategies that will lead to better expected outcomes, the similarity between these two graphs adds further weight to our suggestion that, in fact, the self-reported strategies of players have very little to do with the final outcome.

To answer (Q3) we examine cases in which the blameworthiness score should be zero, and then compare cases that should have lower or higher scores with respect to one another. Once again, comprehensive descriptions of each of our tested queries are omitted for reasons of space, but here we present some representative examples.\footnote{In all of the blameworthiness scores below we use the cost importance measure $N = 1$.} Firstly, we considered level 1 of the game by choosing an alternative distribution $\Pr'$ over contexts when generating our scores. Here a player is less likely to receive a low rating for quality ($Q_1$ or $Q_2$) if they employ a skill-based strategy where tasks are more frequently allocated to better workers ($S$). As expected, our system returns $db_N(S, \neg S, Q_1 \lor Q_2) = 0$. Secondly, we look at the timeliness outcomes. A player is less likely to obtain the top timeliness rating ($T_5$) if they do \textit{not} use a strategy that uniformly allocates tasks ($U$) compared to their \textit{not} using a random strategy of allocation ($R$). Accordingly, we find that $db_N(\neg U, \neg T_5) > db_N(\neg R, \neg T_5)$, and more specifically we have $db_N(\neg U, \neg T_5) = 0.002$ and $db_N(\neg R, \neg T_5) = 0$ (i.e. a player should avoid using a random strategy completely if they wish to obtain the top timeliness rating).

\section{Datasets}
\label{data}

The full set of data, source code, and other supplementary materials can be found online \cite{code}. Here we provide brief summaries of the three datasets used in our experiments, including the variable encoding used for each domain and the underlying constraints.

\begin{table}[H]
	\centering
	\begin{tabular}{p{0.3\linewidth} p{0.6\linewidth}}
	\toprule
		Number of data points & 100000\\
		Number of variables & 12\\
		Context variables & Mediastinal Metastases ($MM$), CT Would Be Positive ($CT_+$), CT Would Be Negative ($CT_-$), Mediastinoscopy Would Be Positive ($M_+$), Mediastinoscopy Would Be Negative ($M_-$)\\
		Decision variables ($\mathcal{D}$) & Perform CT ($CT$), Perform Mediastinoscopy ($M$) \\
		Outcome variables ($\mathcal{O}$) & No CT Performed ($CT_{N/A}$), No Mediastinoscopy Performed ($M_{N/A}$), Thoractomy Performed ($T$), Diagnosis Procedures Survived ($S_{DP}$), Treatment Survived ($S_T$) \\
		Constraints & $(CT_+ \lor CT_-) \leftrightarrow CT$ \newline 
		$CT_{N/A} \leftrightarrow \neg CT$ \newline 
		$(M_+ \lor M_-) \leftrightarrow M$ \newline 
		$M_{N/A} \leftrightarrow \neg M$ \newline 
		$M_- \rightarrow T$ \newline 
		$M_+ \rightarrow \neg T$ \newline 
		$(CT_- \land \neg M) \rightarrow T$ \newline 
		$(CT_+ \land \neg M) \rightarrow \neg T$ \newline 
		$\neg S_{DP} \rightarrow M$ \newline 
		$\neg(CT_+ \land CT_-)$ \newline 
		$\neg(M_+ \land M_-)$ \newline 
		$\neg S_{DP} \rightarrow \neg S_T$ \\
		Model count & 52\\
		Utilities given? & Yes (life expectancy)\\
	\bottomrule
	\end{tabular}
	\caption{A summary of the lung cancer staging data used in our first experiment.}
	\label{tab:cancerdata}
\end{table}

\begin{table}[H]
	\centering
	\begin{tabular}{p{0.3\linewidth} p{0.6\linewidth}}
	\toprule
		Number of data points & 7446\\
		Number of variables & 21\\
		Context variables ($\mathcal{X}$) & Level 1 ($L_1$), ... , Level 6 ($L_6$) \\
		Decision variables ($\mathcal{D}$) & Other ($O$), Load-balancing ($L$), Uniform ($U$), Skill-based ($S$), Random ($R$)\\
		Outcome variables ($\mathcal{O}$) & Timeliness 1 ($T_1$), ... , Timeliness 5 ($T_5$), Quality 1 ($Q_1$), ... , Quality 5 ($Q_5$) \\
		Constraints & $\bigvee_{i \in \{1,...,6\}}L_i$ \newline 
		$L_i \rightarrow \neg\bigvee_{j \in \{1,...,6\}\setminus i}L_j \forall i \in \{1,..., 6\}$ \newline
		$\bigvee_{i \in \{1,...,5\}}T_i$ \newline 
		$T_i \rightarrow \neg\bigvee_{j \in \{1,...,5\}\setminus i}T_j \forall i \in \{1,..., 5\}$ \newline
		$\bigvee_{i \in \{1,...,5\}}Q_i$ \newline 
		$Q_i \rightarrow \neg\bigvee_{j \in \{1,...,5\}\setminus i}Q_j \forall i \in \{1,..., 5\}$\\
		Model count & 4800\\
		Utilities given? & Yes (self-reported happiness score)\\
	\bottomrule
	\end{tabular}
	\caption{A summary of the teamwork management data used in our second experiment.}
	\label{tab:teamworkdata}
\end{table}

\begin{table}[H]
	\centering
	\begin{tabular}{p{0.3\linewidth} p{0.6\linewidth}}
	\toprule
		Number of data points & 360\\
		Number of variables & 23\\
		Context variables ($\mathcal{X}$) & One Person On Track A ($A_1$), ... , Family On Track A ($A_{Fa}$), One Person On Track B ($B_1$), ... , Family On Track B ($B_{Fa}$) \\
		Decision variables ($\mathcal{D}$) & Inaction ($I$), Flip Switch ($F$), Push B ($P$), Sacrifice Oneself ($S$) \\
		Outcome variables ($\mathcal{O}$) & One Person Lives ($L_1)$, ... , Family Lives ($L_{Fa}$), You Live ($L_Y$) \\
		Constraints & $\bigvee_{i \in \{1,...,Fa\}}A_i$ \newline 
		$\bigvee_{i \in \{1,...,Fa\}}B_i$ \newline 
		$\neg(A_i \land B_i) \forall i \in \{1, ... , Fa\}$ \newline 
		$A_i \rightarrow \neg\bigvee_{j \in \{1,...,Fa\}\setminus i}A_j \forall i \in \{1,..., Fa\}$ \newline 
		$B_i \rightarrow \neg\bigvee_{j \in \{1,...,Fa\}\setminus i}B_j \forall i \in \{1,..., Fa\}$ \newline
		$\bigvee_{D \in \{N,F,P,S\}}D$ \newline 
		$D \rightarrow \neg\bigvee_{D' \in \{N,F,P,S\} \setminus D}D'$ \newline 
		$(A_i \land N) \rightarrow \neg L_i \forall i \in \{1,..., Fa\}$ \newline 
		$(B_i \land N) \rightarrow L_i \forall i \in \{1,..., Fa\}$ \newline 
		$L_i \rightarrow (A_i \lor B_i) \forall i \in \{1,..., Fa\}$ \newline 
		$(S \land (A_i \lor B_i)) \rightarrow L_i \forall i \in \{1,..., Fa\}$ \newline 
		$L_Y \leftrightarrow \neg S$ \newline 
		$(L_i \land (P \lor F)) \rightarrow \neg\bigvee_{j \in \{1,...,Fa\}\setminus i}L_j \forall i \in \{1,..., Fa\}$ \newline 
		$(\neg L_i \land (P \lor F)) \rightarrow \bigvee_{j \in \{1,...,Fa\}\setminus i}L_j \forall i \in \{1,..., Fa\}$ \\
		Model count & 180\\
		Utilities given? & No\\
	\bottomrule
	\end{tabular}
	\caption{A summary of the trolley problem data used in our third experiment.}
	\label{tab:trolleydata}
\end{table}

\end{document}